%% file: emnlp_main.tex
\documentclass[11pt]{article}

\usepackage[preprint]{acl}

\usepackage{times}
\usepackage{latexsym}

\usepackage[T1]{fontenc}

\usepackage[utf8]{inputenc}

\usepackage{microtype}

\usepackage{inconsolata}

\usepackage{algorithm}
\usepackage{algpseudocode}
\usepackage{booktabs}
\usepackage{multirow}
\usepackage{graphicx}
\usepackage{enumitem}
\usepackage{tcolorbox}
\usepackage{amsmath}
\usepackage{amssymb}

\newcommand{\cmark}{$\checkmark$}

\newtheorem{proposition}{Proposition}

\newenvironment{justification}{%
  \par\noindent\textit{Justification.}\space\ignorespaces
}{%
  \hfill$\square$\par\medskip
}

\title{All Leaks Count, Some Count More: Interpretable Temporal Contamination Detection and Mitigation in LLM Backtesting}

\author{
  \textbf{Zeyu Zhang\textsuperscript{1}},
  \textbf{Ryan Chen\textsuperscript{1}},
  \textbf{Bradly C. Stadie\textsuperscript{1,2}}
\\
\\
  \textsuperscript{1}Department of Statistics and Data Science, Northwestern University
\\
  \textsuperscript{2}Bridgewater AIA Labs
\\
  \small{
    \textbf{Correspondence:} \href{mailto:zeyuzhang2028@u.northwestern.edu}{zeyuzhang2028@u.northwestern.edu}
  }
}

\begin{document}
\maketitle

\begin{abstract}
  Backtesting LLMs on resolved events assumes models reason only from pre-cutoff knowledge, yet pretrained models inevitably leak post-cutoff knowledge. We introduce a claim-level evaluation framework that decomposes prediction rationales into atomic claims and applies Shapley values to quantify each claim's decision impact, yielding \textbf{Shapley-DCLR} (\textbf{Shapley}-weighted \textbf{D}ecision-\textbf{C}ritical \textbf{L}eakage \textbf{R}ate)---an interpretable metric measuring what fraction of decision-driving reasoning is contaminated. We further propose \textbf{TimeSPEC} (\textbf{Time}-\textbf{S}upervised \textbf{P}rediction with \textbf{E}xtracted \textbf{C}laims), an inference-time architecture that interleaves temporally-filtered retrieval with claim-level supervision, producing predictions grounded entirely in pre-cutoff evidence. Across three LLMs, the ablation experiments confirm retrieval and supervision are jointly necessary; and a three-task probe further illstrates that the performance cost of temporal enforcement scales with each task's reliance on post-cutoff information.
\end{abstract}


\input{sections/emnlp_introduction}
\input{sections/emnlp_methodology}
\input{sections/emnlp_temporal_llm_agent}
\input{sections/emnlp_experiments}
\input{sections/emnlp_related_work}
\input{sections/emnlp_conclusion}

\section*{Limitations}

TimeSPEC operates at inference time, so leaked information persists in parametric memory and is \emph{blocked} rather than \emph{removed}; temporally partitioned pretraining or reinforcement learning that teaches temporal discipline may address leakage at the source. The evaluation framework incurs per-instance cost from Shapley coalition evaluations and search-based verification; Leverage SHAP reduces the Shapley budget to $O(n\log n)$ but the overhead remains non-trivial. The TimeSPEC generation pipeline adds further cost through multi-phase supervision and regeneration. Finally, while our experiments span binary forecasting, numerical estimation, and ranking across diverse domains, extending the framework to broader question types---particularly open-ended generation---remains future work.



\bibliography{emnlp_references}

\appendix
\input{sections/emnlp_appendix}

\end{document}

%% file: sections/emnlp_introduction.tex

\section{Introduction}
\label{sec:introduction}

LLMs have demonstrated strong prediction capabilities across domains from legal outcomes to financial metrics~\citep{brown2020language,wu2023bloomberggpt}, raising the question of whether they can accuratelyforecast \emph{future} events. Answering this requires \emph{backtesting}---evaluating on historical events where ground truth is known~\citep{bailey2014determining}---under the assumption that models reason only from information available at a reference date $t_{\text{ref}}$.

Models pretrained on web-scale corpora inevitably encode events after $t_{\text{ref}}$, so an LLM may ``recall'' outcomes from training data, producing accurate but illegitimate predictions---a problem termed \emph{temporal knowledge leakage}~\citep{iclr2026forecasting,sarkar2025lookahead}. Retrieval compounds the risk: even with date filters, post-cutoff content passes through unreliable metadata~\citep{ellahib2026datefilter}. Recent efforts---train-time temporal partitioning~\citep{drinkall2024timemachine,he2025chronogpt}, benchmark contamination audits~\citep{forecastbench2025}, and memorization detection~\citep{carlini2021extracting,jiang2024contamination}---mitigate aspects of this problem, yet none provides fine-grained attribution of \emph{which} information constitutes leakage and \emph{how much} it influences predictions.

We address these gaps with two contributions. First, we introduce a model-agnostic, claim-level evaluation framework: a model's prediction rationale is decomposed into atomic claims, each grounded to a temporal provenance. Rather than binary leaked-or-clean labels, we apply Shapley values~\citep{shapley1953value,lundberg2017unified} to quantify each claim's marginal contribution to the prediction, yielding \textbf{Shapley-DCLR} (\textbf{Shapley}-weighted \textbf{D}ecision-\textbf{C}ritical \textbf{L}eakage \textbf{R}ate)---an interpretable metric that captures what fraction of decision-driving reasoning derives from leaked information. Second, we propose \textbf{TimeSPEC} (\textbf{Time}-\textbf{S}upervised \textbf{P}rediction with \textbf{E}xtracted \textbf{C}laims), an inference-time architecture that proactively prevents leakage. TimeSPEC interleaves generation with claim-level supervision; when temporal violations are detected, regeneration produces corrected output using only validated evidence. Unlike prompt-based temporal constraints---which fail because LLMs cannot reliably self-assess the temporal validity of their own knowledge---TimeSPEC enforces validity through programmatic supervision and a closed-world aggregator that admits only verified claims.

We evaluate on 2{,}769 binary forecasting instances across three LLMs and three prediction domains, applying Shapley-DCLR to every baseline. A $2{\times}2$ ablation over (Search, Supervision) reveals that standard prompting and unfiltered retrieval both exhibit substantial leakage, and establishes TimeSPEC's two mechanisms as jointly necessary: neither alone matches the joint objective of low leakage and retained accuracy. A complementary three-task probe---Stock ranking, NBA Salary, and Legal---shows that the accuracy cost of strict temporal enforcement tracks each task's intrinsic reliance on post-cutoff information, serving as a per-task diagnostic of how much unfiltered accuracy was contamination.

%% file: sections/emnlp_methodology.tex

\section{Temporal Leakage Evaluation Framework}
\label{sec:eval_framework}

We introduce a framework for auditing temporal leakage in LLM rationales. The framework is \emph{model-agnostic}, consuming only the rationale and the reference date $t_{\text{ref}}$ so that it applies to any LLM without access to model internals. It is \emph{fine-grained}: every atomic claim in the rationale receives an independent leakage decision, in contrast to instance-level contamination metrics that flag whole items. And it is \emph{interpretable}: each leaked claim carries a Shapley-derived weight that quantifies its contribution to the prediction, exposing \emph{which} statements move the decision and \emph{how much}.

\subsection{Problem Setup and Temporal Leakage}
\label{subsec:problem_setup}

Consider a prediction task in which an LLM $\mathcal{M}$ forecasts the outcome of an event $E$ resolved at time $t_E$, conditioned on a reference time $t_{\text{ref}} < t_E$. Backtesting requires that $\mathcal{M}$ reason only from information available before $t_{\text{ref}}$. Let $\mathcal{K}(t)$ denote the cumulative public knowledge at time $t$, satisfying monotonicity: $t_1 \leq t_2 \implies \mathcal{K}(t_1) \subseteq \mathcal{K}(t_2)$. The model produces a rationale $R$ and a prediction $\hat{y}$. We ask whether $R$ derives solely from $\mathcal{K}(t_{\text{ref}})$ or incorporates information from $\mathcal{K}(t) \setminus \mathcal{K}(t_{\text{ref}})$ for some $t > t_{\text{ref}}$.

For fine-grained analysis, we decompose the rationale into atomic claims $R = \{c_1, \ldots, c_n\}$, each a single verifiable assertion. The \emph{temporal grounding function}
\begin{equation}
    \tau(c) = \inf \bigl\{ t : c \in \mathcal{K}(t) \;\text{or}\; c \text{ derivable from } \mathcal{K}(t) \bigr\}
\label{eq:temporal_grounding}
\end{equation}
returns the earliest time at which $c$ is publicly knowable, with $\tau(c) = -\infty$ for timeless truths. A claim exhibits \emph{temporal leakage} iff $\tau(c) > t_{\text{ref}}$, recorded by the indicator $\ell(c; t_{\text{ref}}) = \mathbf{1}[\tau(c) > t_{\text{ref}}]$.

\subsection{Shapley-Value Decision Weighting}
\label{subsec:shapley_weighting}

Per-claim leakage is necessary but not sufficient: a leaked claim that does not move the prediction is a less serious failure than one that does. We therefore weight each claim by its contribution to the prediction using Shapley values.

\paragraph{Shapley-value attribution.}
Let $N = \{1, \ldots, n\}$ index the claims of a rationale and let $v: 2^N \to \mathbb{R}$ be a characteristic function with $v(S)$ equal to the normalized above-baseline prediction score when only claims in subset $S$ are visible to $\mathcal{M}$. The \emph{Shapley value} of claim $c_i$ is
\begin{equation}
\phi_i = \sum_{S \subseteq N \setminus \{i\}} \frac{|S|!\,(n{-}|S|{-}1)!}{n!} \bigl[ v(S \cup \{i\}) - v(S) \bigr],
\label{eq:shapley}
\end{equation}
the expected marginal contribution of $c_i$ across all orderings in which claims could enter the rationale. The Shapley framework~\citep{shapley1953value,lundberg2017unified} maps cleanly onto our setting: \emph{players} are atomic claims, \emph{coalitions} are claim subsets, and $v(S)$ is the prediction score when the rationale is restricted to $S$. Shapley values are the unique axiomatic attribution under this characteristic function: they partition the rationale's net contribution into per-claim shares ($\sum_i \phi_i = v(N) - v(\emptyset)$) and assign zero weight to any claim whose marginal contribution to every coalition is zero.

\paragraph{Tractability.}
Exact computation of $\boldsymbol{\phi}$ requires $2^n$ coalition evaluations, intractable for typical rationales with 10--20+ claims. Classical Monte Carlo permutation sampling is unbiased but converges only at $\mathrm{Var}(\hat{\phi}_i) = O(1/m)$ without a high-probability precision guarantee. We adopt Leverage SHAP~\citep{musco2025leverage}, which reframes Shapley estimation as leverage-weighted regression and admits a provably tight sample complexity:
\begin{equation}
m = O(C \cdot n \cdot \log n)
\label{eq:leverage_budget}
\end{equation}
coalitions suffice for a $(1{+}\epsilon)$-approximation to $\boldsymbol{\phi}$ with high probability, where the constant $C$ controls precision. The exponential budget of exact computation is thus reduced to near-linear in $n$ while the axiomatic properties of Eq.~\ref{eq:shapley} are preserved.

\subsection{Leakage Evaluation Pipeline}
\label{subsec:eval_pipeline}

\begin{figure*}[t]
\centering
\includegraphics[width=0.90\textwidth]{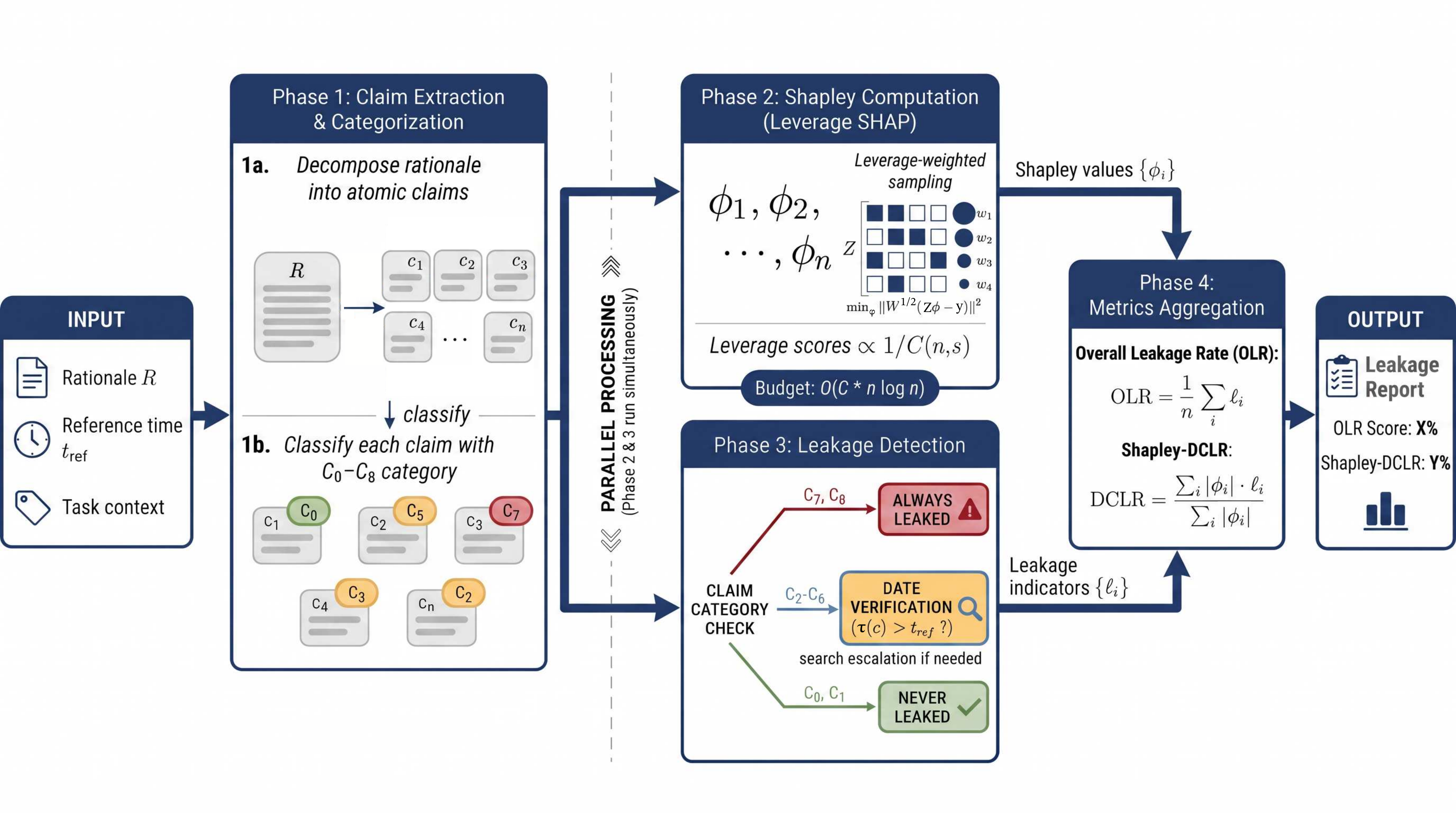}
\caption{\textbf{The Shapley-DCLR evaluation pipeline: a model-agnostic, claim-level framework for detecting and quantifying temporal leakage.} Given any LLM's rationale $R$ and a reference date $t_{\text{ref}}$, \textbf{Phase~1} decomposes $R$ into atomic claims and categorizes each. \textbf{Phases~2 and~3} run in parallel. Phase~2 estimates each claim's Shapley contribution $\phi_i$ to the prediction via Leverage SHAP at $O(C\!\cdot\!n\log n)$ cost. Phase~3 determines leakage indicators $\{\ell_i\}$ via category-based routing, where deterministic categories bypass external search, reducing verification cost. \textbf{Phase~4} fuses $\{\phi_i\}$ and $\{\ell_i\}$ into Shapley-DCLR (Eq.~\ref{eq:shapley_dclr}).}
\label{fig:evaluation_pipeline}
\end{figure*}

Given a rationale $R$ and reference date $t_{\text{ref}}$, the framework runs the four-phase pipeline illustrated in Figure~\ref{fig:evaluation_pipeline}; Phases~2 and~3 execute in parallel.

\paragraph{Phase 1: Claim extraction and categorization.}
The extractor decomposes $R$ into atomic claims $\{(c_i, \kappa_i)\}_{i=1}^n$, where each $c_i$ is a single verifiable assertion satisfying atomicity, self-containment, factuality, and temporal grounding, and $\kappa_i \in \{\text{C0},\ldots,\text{C8}\}$ is its category under a nine-class taxonomy. The taxonomy partitions claims into three operational tiers---\emph{deterministically safe} (C0--C1), \emph{requires date verification} (C2--C6), and \emph{deterministically leaked} (C7--C8)---enabling the category-based routing in Phase~3. Full specification and worked examples are in Appendix~\ref{app:taxonomy_full}.

\paragraph{Phase 2: Shapley computation.}
Phase~2 instantiates the characteristic function $v$ of Section~\ref{subsec:shapley_weighting} by re-running $\mathcal{M}$ on coalitions sampled via Leverage SHAP (Eq.~\ref{eq:leverage_budget}), producing per-claim Shapley values $\{\phi_i\}$ at near-linear cost.

\paragraph{Phase 3: Leakage detection.}
Category-based routing resolves each claim's leakage indicator $\ell_i$. Claims whose leakage status is structurally determined are resolved without search. For the remaining claims, the pipeline queries external sources to establish $\tau(c)$ and evaluates $\tau(c) > t_{\text{ref}}$. The authority is the externally-verified publication date of multiple retrieved sources, which is more reliable than model-declared or in-claim dates that can be hallucinated or ambiguous. Vague temporal references are resolved conservatively (e.g., ``2023'' maps to December~31, 2023) to avoid false negatives from under-specified timestamps.

\paragraph{Phase 4: Metrics aggregation.}
Phase~4 fuses the Shapley values $\{\phi_i\}$ from Phase~2 with the leakage indicators $\{\ell_i\}$ from Phase~3 into the \emph{Shapley-weighted Decision-Critical Leakage Rate}:
\begin{equation}
\text{Shapley-DCLR} = \frac{\sum_{i=1}^{n} |\phi_i| \cdot \ell(c_i)}{\sum_{i=1}^{n} |\phi_i|},
\label{eq:shapley_dclr}
\end{equation}
which admits two complementary readings. As \textit{influence-weighted leakage}, it measures whether the claims that actually drive the prediction are contaminated---high Shapley-DCLR means decision-critical claims are leaked while a low value indicates peripheral leakage. As \textit{weighted information composition}, it reports the fraction of decision-relevant evidence that derives from post-cutoff sources, enabling both instance-level diagnosis and aggregate quality assessment. 

%% file: sections/emnlp_temporal_llm_agent.tex

\begin{figure*}[t]
    \centering
    \includegraphics[width=0.80\textwidth]{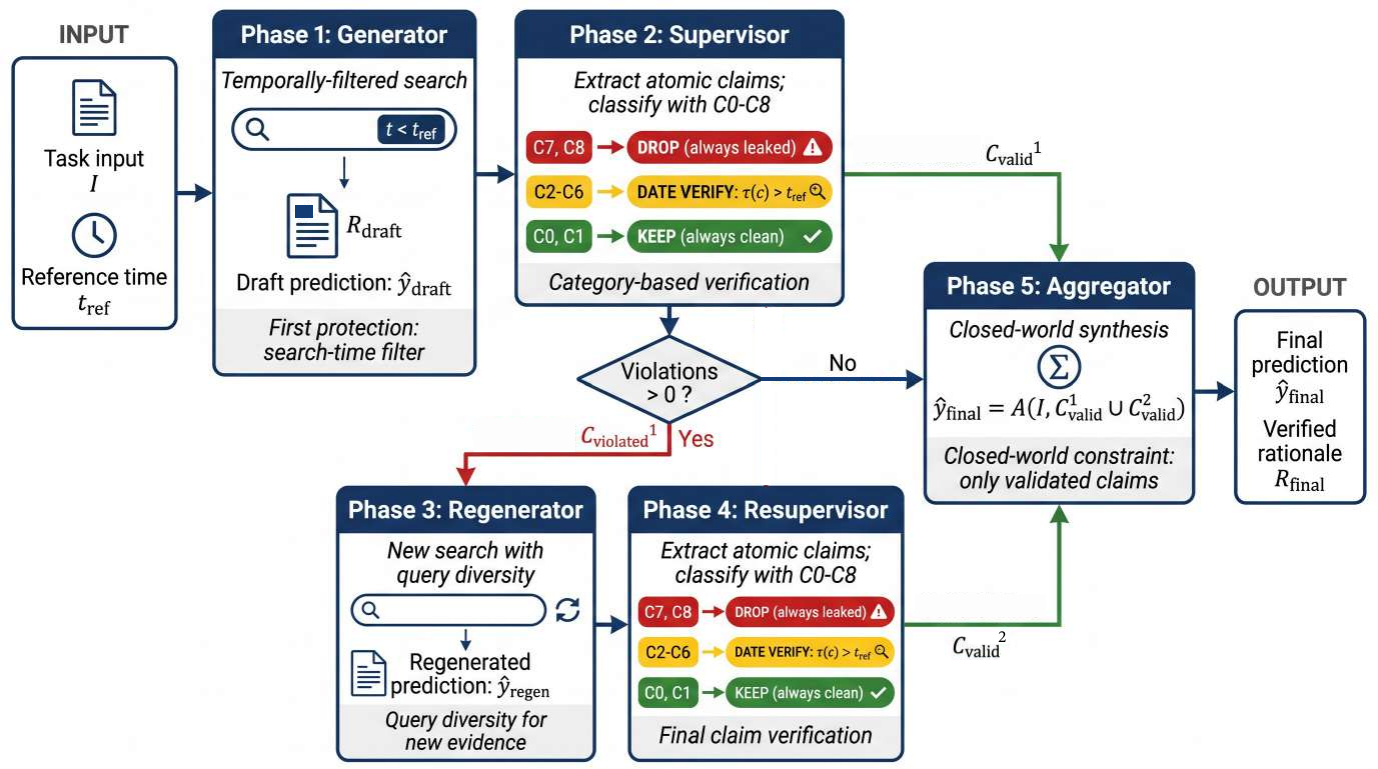}
    \caption{\textbf{The TimeSPEC architecture: an inference-time pipeline that interleaves temporally-filtered retrieval with claim-level supervision under a closed-world guarantee.} The \textbf{Generator} issues date-filtered web queries ($t < t_{\text{ref}}$) and produces a draft prediction with rationale. The \textbf{Supervisor} extracts and categorizes claims, partitioning them into validated ($\mathcal{C}_{\text{valid}}^{(1)}$) and violated sets. If violations exist, the \textbf{Regenerator} re-queries with diversified searches and the \textbf{Resupervisor} audits the new rationale, yielding $\mathcal{C}_{\text{valid}}^{(2)}$; if none, the validated claims route directly to the \textbf{Aggregator}. The Aggregator synthesizes $\hat{y}_{\text{final}} = \mathcal{A}(\mathcal{I},\, \mathcal{C}_{\text{valid}}^{(1)} \cup \mathcal{C}_{\text{valid}}^{(2)})$ under a closed-world constraint: only the task input and validated claims enter the final synthesis.}
    \label{fig:timespec}
\end{figure*}

\section{TimeSPEC: Inference-Time Forecasting Architecture}
\label{sec:timespec}

TimeSPEC is an inference-time forecasting architecture combining two complementary mechanisms: temporally-filtered retrieval that supplies date-filtered pre-cutoff evidence, and claim-level supervision that audits every generated claim for temporal violations. When violations are detected, a regeneration loop re-queries diverse sources to replace leaked content. A closed-world aggregator then synthesizes the final prediction from only the task input and validated claims, structurally excluding unaudited parametric content. The five phases are shown in Figure~\ref{fig:timespec}.

\paragraph{Phase~1: Generator.}
Given task input $\mathcal{I}$ and reference date $t_{\text{ref}}$, the Generator issues temporally-filtered web queries with date constraints ($t < t_{\text{ref}}$) and produces a draft prediction $\hat{y}_{\text{draft}}$ with rationale $R_{\text{draft}}$. API-level date filtering provides the first protection layer but is insufficient on its own: date-restricted retrieval still admits post-cutoff content through unreliable metadata~\citep{ellahib2026datefilter}, and no retrieval constraint can block the LLM's parametric memory from re-introducing post-cutoff facts at generation time.

\paragraph{Phase~2: Supervisor.}
To address the limitations of API-level filtering, the Supervisor extracts atomic claims from $R_{\text{draft}}$, categorizes them, and determines each claim's leakage status via category-based routing. Claims are partitioned into a validated set $\mathcal{C}_{\text{valid}}^{(1)}$ and a violated set $\mathcal{C}_{\text{violated}}^{(1)}$. If no violations are found, $\mathcal{C}_{\text{valid}}^{(1)}$ routes directly to Phase~5; otherwise regeneration is triggered.

\paragraph{Phase~3: Regenerator.}
Activated when violations exist, the Regenerator inherits the validated claims $\mathcal{C}_{\text{valid}}^{(1)}$ and the prior search queries, then issues new queries constrained to be diverse from those already attempted, encouraging retrieval to unexplored pre-cutoff sources. The phase produces an updated prediction $\hat{y}_{\text{regen}}$ with rationale $R_{\text{regen}}$. 

\paragraph{Phase~4: Resupervisor.}
The Resupervisor applies the same claim-level verification as Phase~2 to $R_{\text{regen}}$, producing $\mathcal{C}_{\text{valid}}^{(2)}$. Any residual violations are excluded from aggregation and logged for transparency.

\paragraph{Phase~5: Aggregator.}
The Aggregator synthesizes the final prediction from the task input and the union of all validated claims:
\begin{equation}
\hat{y}_{\text{final}} = \mathcal{A}\bigl(\mathcal{I},\; \mathcal{C}_{\text{valid}}^{(1)} \cup \mathcal{C}_{\text{valid}}^{(2)}\bigr),
\label{eq:aggregation}
\end{equation}
under a \emph{closed-world} constraint: only $\mathcal{I}$ and the validated claims enter the synthesis prompt, so no unaudited parametric content or unverified retrieved information can influence the final answer. The output is both a prediction $\hat{y}_{\text{final}}$ and a verified rationale $R_{\text{final}}$ grounded entirely in dated, categorized evidence.

%% file: sections/emnlp_experiments.tex

\begin{table*}[t]
    \centering
    \caption{Treatment-group results across the (Search, Supervision) ablation on three LLMs. DCLR: Shapley-DCLR (\%, $\downarrow$, lower is less leakage); Acc: $1{-}\text{Brier}$ (\%, $\uparrow$). Temporal RAG achieves the highest accuracy but also the highest leakage on every model; TimeSPEC recovers most of the accuracy gain while suppressing DCLR to the supervised floor. Bold marks the best value per column.}
    \label{tab:main_treatment}
    \footnotesize
    \setlength{\tabcolsep}{4pt}
    \begin{tabular}{@{}lcccccccc@{}}
    \toprule
    &  &  & \multicolumn{2}{c}{Qwen3.5-35B} & \multicolumn{2}{c}{Kimi-K2.5} & \multicolumn{2}{c}{Claude Sonnet 4} \\
    \cmidrule(lr){4-5} \cmidrule(lr){6-7} \cmidrule(lr){8-9}
    Method & Search & Sup. & DCLR\,(\%)$\downarrow$ & Acc\,(\%)$\uparrow$ & DCLR\,(\%)$\downarrow$ & Acc\,(\%)$\uparrow$ & DCLR\,(\%)$\downarrow$ & Acc\,(\%)$\uparrow$ \\
    \midrule
    Temporal Hint    & --      & --      & 5.2           & 80.5          & 2.7           & 86.8          & 1.0           & 83.9          \\
    Self-Correction  & --      & \cmark  & \textbf{1.6}  & 79.9          & 1.7           & 86.8          & \textbf{0.8}  & 84.0          \\
    Temporal RAG     & \cmark  & --      & 13.1          & \textbf{85.4} & 8.9           & \textbf{87.8} & 4.7           & \textbf{84.2} \\
    TimeSPEC         & \cmark  & \cmark  & 1.8           & 83.9          & \textbf{1.0}  & 87.4          & \textbf{0.8}  & 83.9          \\
    \bottomrule
    \end{tabular}
\end{table*}

\begin{table}[t]
    \centering
    \caption{Control-group Shapley-DCLR (\%, $\downarrow$) on three LLMs. Control questions resolve after every model's knowledge cutoff, so parametric leakage is impossible by construction. Temporal Hint and Self-Correction confirm the near-zero empirical floor.}
    \label{tab:main_control}
    \footnotesize
    \setlength{\tabcolsep}{4pt}
    \begin{tabular}{@{}lccc@{}}
    \toprule
    Method & Qwen3.5 & Kimi-K2.5 & Claude \\
    \midrule
    Temporal Hint    & 0.6 & 0.5 & 0.2 \\
    Self-Correction  & 0.5 & 0.4 & 0.2 \\
    Temporal RAG     & 16.7 & 17.0 & 18.5 \\
    TimeSPEC         & 3.7 & 2.7 & 4.2 \\
    \bottomrule
    \end{tabular}
\end{table}

\section{Experiments}
\label{sec:experiments}

The preceding sections introduced two contributions: a claim-level evaluation framework centered on Shapley-DCLR (Section~\ref{sec:eval_framework}), and TimeSPEC, which combines temporally-filtered retrieval with claim-level supervision (Section~\ref{sec:timespec}). We now validate both empirically. An ablation over (Search, Supervision) on 2{,}769 instances across three LLMs tests whether the two mechanisms are jointly necessary and, via a Treatment/Control split, validates the metric (Section~\ref{subsec:main_results}). A complementary three-task probe---Stock ranking, NBA Salary, and Legal---shows that the performance effect of strict temporal enforcement tracks each task's intrinsic reliance on post-cutoff information, serving as a per-task diagnostic of how much unfiltered performance was contamination (Section~\ref{subsec:analysis}).

\paragraph{Shared setup.}
Both experiments use four baselines occupying the cells of a (Search, Supervision) grid. \textbf{Temporal Hint} ($-$Search, $-$Supervision) issues only a prompt-level date constraint. \textbf{Self-Correction} ($-$Search, $+$Supervision) adds self-review. \textbf{Temporal RAG} ($+$Search, $-$Supervision) grants date-filtered retrieval but no claim-level verification. \textbf{TimeSPEC} ($+$Search, $+$Supervision) is the full pipeline of Section~\ref{sec:timespec}. Leakage is reported as Shapley-DCLR (Eq.~\ref{eq:shapley_dclr}) computed via Leverage SHAP; accuracy metrics are experiment-specific. Full implementation details are in Appendix~\ref{app:implementations}; code is available at \url{https://anonymous.4open.science/r/TimeSPEC-6E24}.

\subsection{Large-Scale Ablation}
\label{subsec:main_results}

\paragraph{Data and models.}
The evaluation comprises 2{,}769 binary forecasting questions partitioned into a Treatment group (1{,}942 questions from Autocast~\citep{zou2022forecasting}, all resolved before every model's knowledge cutoff) and a Control group (827 questions from ForecastBench~\citep{forecastbench2025}, all resolving after every model's cutoff, so parametric leakage is ruled out by construction). Three LLMs span distinct vendors and knowledge cutoffs: Qwen3.5-35B~\citep{qwen2026qwen35}, Kimi-K2.5~\citep{team2026kimik25}, and Claude Sonnet~4~\citep{anthropic2025claude}. Accuracy is $1{-}\mathrm{Brier}$, where $\mathrm{Brier} = (\hat{p} - y)^2$, $\hat{p} \in [0,1]$ is the predicted probability and $y \in \{0,1\}$ is the outcome; higher is better. Dataset details appear in Appendix~\ref{app:datasets}.

\paragraph{Shapley-DCLR is a valid and responsive metric.}
The Control group establishes the empirical floor: Temporal Hint and Self-Correction produce near-zero DCLR on every model ($0.6\%/0.5\%$ Qwen3.5, $0.5\%/0.4\%$ Kimi, $0.2\%/0.2\%$ Claude), confirming that the metric does not flag leakage-free predictions as contaminated. The small residual reflects occasional hallucinated claims that happen to carry post-cutoff dates (e.g., fabricating a future policy announcement), not genuine leakage. Comparing Treatment to Control at Temporal Hint reveals the metric's sensitivity: DCLR rises $8.7\times$ on Qwen3.5 ($5.2\%$ vs.\ $0.6\%$), $5.4\times$ on Kimi, and $5.0\times$ on Claude once parametric leakage becomes possible. Together, the near-zero Control floor and the several-fold Treatment elevation validate Shapley-DCLR as both precise (few false positives) and sensitive to genuine contamination.

\paragraph{Retrieval improves accuracy but amplifies leakage.}
Even without search, Temporal Hint already exhibits non-trivial parametric leakage (Treatment DCLR $5.2\%/2.7\%/1.0\%$), confirming that models encode post-cutoff knowledge in their parameters. Adding date-filtered retrieval (Temporal RAG) improves accuracy on every model ($+4.9/+1.0/+0.3$ pts) but amplifies DCLR by $2.5$--$4.7\times$ ($5.2\%\to13.1\%$ Qwen3.5, $2.7\%\to8.9\%$ Kimi, $1.0\%\to4.7\%$ Claude): contamination scales faster than genuine signal. The Control group isolates the retrieval channel directly: with parametric leakage ruled out, Temporal RAG still produces DCLR of $16.7\%/17.0\%/18.5\%$, so every leaked claim must originate from retrieved documents. A hard date filter at the search API is therefore insufficient---post-cutoff knowledge propagates through unreliable publication-date metadata and post-publication page edits~\citep{ellahib2026datefilter}. Retrieval adds genuine predictive signal, but without claim-level verification it simultaneously introduces a contamination channel that exceeds the parametric baseline.

\paragraph{Both mechanisms in TimeSPEC are jointly necessary.}
Two pairwise comparisons test whether each mechanism remains non-redundant in the presence of the other. Adding search to Self-Correction (Self-Correction $\to$ TimeSPEC) recovers accuracy by up to $+4.0$ pts on Qwen3.5 ($79.9\%\to83.9\%$) while DCLR stays at the supervised floor ($1.6\%\to1.8\%$): the supervisor catches the contamination retrieval introduces, so the accuracy gain is decoupled from a leakage gain. Adding supervision to Temporal RAG (Temporal RAG $\to$ TimeSPEC) cuts DCLR by $83$--$89\%$ on every model ($13.1\%\to1.8\%$, $8.9\%\to1.0\%$, $4.7\%\to0.8\%$) at bounded accuracy cost ($-1.5$, $-0.4$, $-0.3$ pts). Neither mechanism alone matches TimeSPEC: Self-Correction lacks the evidence retrieval supplies, and Temporal RAG lacks the filter that strips retrieval-borne contamination. The pattern holds on all three models, confirming that the joint-necessity conclusion is not specific to one vendor's training data or alignment recipe.

\subsection{TimeSPEC under Varying Leakage Sensitivity}
\label{subsec:analysis}

The large-scale ablation validates TimeSPEC's two mechanisms as jointly necessary on average. A natural follow-up is whether the framework honestly reflects a model's true prediction ability on individual tasks---particularly those where temporally valid information alone is insufficient for accurate prediction. On such tasks, models may silently draw on memorized or retrieved post-cutoff knowledge, inflating reported metrics and undermining backtesting integrity. We probe this by constructing three forecasting tasks that deliberately span the leakage-sensitivity spectrum: Stock ranking (high), NBA Salary (intermediate), and Legal (low). For each task, we measure how leakage suppression affects both DCLR and task-specific performance, and analyze what the resulting changes reveal about each task's reliance on post-cutoff knowledge.

\paragraph{Task design.}
We select three forecasting tasks whose reliance on post-cutoff information varies by construction. \emph{Stock return ranking} is the most leakage-sensitive: market returns are highly volatile and difficult to predict from pre-cutoff fundamentals alone, so models with access to post-cutoff knowledge gain a disproportionate advantage. \emph{NBA Salary prediction} is intermediate: contract values depend on recent player performance and market conditions that may fall within days or weeks of the cutoff, making up-to-date knowledge informative but not as decisive as in stock markets. \emph{Legal case outcome prediction} is least susceptible: judicial outcomes rely primarily on precedent, statutory reasoning, and procedural history available well before the decision, and similar prior cases provide strong valid reference points.

\paragraph{Data, model, and metrics.}
\emph{Stock}: $n{=}100$ ranking instances from the COVID-19 market period (December~2019--June~2020), each containing 5 S\&P~500 stocks from a single sector; $t_{\text{ref}}$ is set to December~1, 2019 (period start). Performance is $(\rho+1)/2$ ($\uparrow$), where $\rho$ is Spearman rank correlation~\citep{spearman1904proof}. \emph{NBA Salary}: $n{=}137$ notable free-agent contracts spanning 2019--2025; the task is to predict each player's annual average value (AAV) in USD, with $t_{\text{ref}}$ set before each contract announcement. Performance is $1 - |\hat{y}-y|/|y|$ ($\uparrow$). \emph{Legal}: $n{=}100$ U.S.\ Supreme Court cases from recent terms; the task is binary classification of whether the petitioner prevails, with $t_{\text{ref}}$ set between oral argument and decision date. Performance is $1 - (\hat{p}-y)^2$ ($\uparrow$), where $\hat{p} \in [0,1]$ is the predicted probability and $y \in \{0,1\}$ the outcome. All baselines run with Qwen3.5-35B. Details are in Appendix~\ref{app:data_probe}.

\begin{figure*}[t]
\centering
\includegraphics[width=0.90\textwidth]{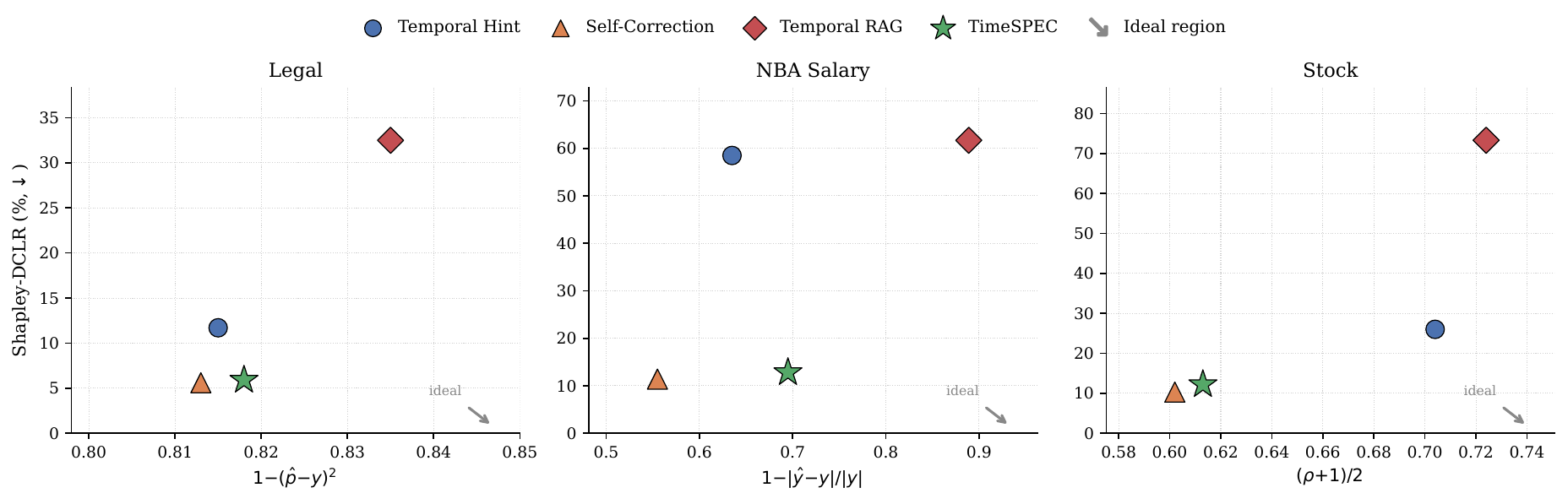}
\caption{Task-level performance vs.\ Shapley-DCLR across the four baselines on Qwen3.5-35B. Each panel plots task-specific performance ($\uparrow$) against Shapley-DCLR (\%, $\downarrow$); lower-right is the ideal operating region. Performance metrics are $1{-}(\hat{p}{-}y)^2$ for Legal, $1{-}|\hat{y}{-}y|/|y|$ for NBA Salary, and $(\rho{+}1)/2$ for Stock, all normalized to $[0,1]$.}
\label{fig:tradeoff}
\end{figure*}

\paragraph{Stock: performance honestly degrades when leakage suppressed}
COVID-era returns are driven by events no December-2019 forecaster could anticipate; any model that ranks these stocks well is likely drawing on post-cutoff knowledge. Self-Correction already drops performance sharply relative to Temporal Hint ($0.602$ vs.\ $0.704$, DCLR $10.3\%$ vs.\ $26.0\%$), confirming that the model's memorized COVID knowledge---not pre-cutoff fundamentals---was the primary source of apparent ranking skill. TimeSPEC reaches $0.613$ at DCLR $12.2\%$, a $0.091$-point drop from Temporal Hint that measures how much of the unfiltered baseline's ranking ability was parametric contamination. Temporal RAG pushes performance to $0.724$ but at DCLR $73.3\%$---nearly three-quarters of its decision-critical reasoning is leaked---making it the clearest illustration of inflated backtesting. What TimeSPEC reports is what an honest pre-COVID analyst could have produced; the performance drop is not a cost but a correction.

\paragraph{NBA Salary: leakage suppressed, estimation improved.}
Salary estimation mixes pre-cutoff signal---player statistics, age, cap structure, comparable prior extensions---with post-cutoff targets: the actual signed contracts that an unfiltered model can memorize. Temporal Hint already carries heavy parametric leakage (DCLR $58.5\%$), as widely-publicized contract values encode directly into model parameters. TimeSPEC strictly dominates Temporal Hint: it reduces DCLR to $12.8\%$ while simultaneously improving estimation accuracy from $0.635$ to $0.695$, because verified pre-cutoff comparables substitute for leaked contract values. Temporal RAG achieves the highest accuracy ($0.889$) but at DCLR $61.7\%$; the $0.194$-point gap to TimeSPEC is the contamination premium---most of Temporal RAG's apparent estimation skill rests on leaked post-cutoff contracts, not genuine pre-cutoff reasoning. On a task where pre-cutoff evidence is informative but incomplete, TimeSPEC demonstrates that honest retrieval outperforms leaked memorization.

\paragraph{Legal: leakage suppressed, prediction preserved.}
Judicial outcomes rely on precedent, statutory text, and prior opinions---an evidentiary base mostly pre-cutoff---so the model has little need to draw on post-cutoff knowledge. All four baselines cluster in a tight performance band ($0.813$--$0.835$), and Self-Correction barely differs from Temporal Hint ($0.813$ vs.\ $0.815$), confirming that parametric leakage contributes negligibly to prediction on this task. Temporal RAG raises DCLR to $32.5\%$ for only a marginal performance gain ($0.835$ vs.\ $0.815$), showing that even substantial leaked knowledge adds little when valid evidence already suffices. TimeSPEC matches Temporal Hint ($0.818$ vs.\ $0.815$) at less than half the DCLR ($5.9\%$ vs.\ $11.7\%$). On a task where temporally valid information suffices, enforcement is effectively free.

\paragraph{The performance effect scales with leakage sensitivity.}
The three tasks reveal a systematic pattern. The parametric leakage benefit---measured by the Temporal Hint-to-Self-Correction performance drop---grows monotonically with task difficulty: smallest on Legal ($-0.002$), intermediate on NBA Salary ($-0.080$), and largest on Stock ($-0.102$). The harder it is to predict from pre-cutoff evidence alone, the more the model silently relies on memorized post-cutoff knowledge. Comparing Temporal RAG to TimeSPEC isolates the cost of honest supervision over unfiltered retrieval: DCLR drops drastically on every task ($32.5\%{\to}5.9\%$ Legal, $61.7\%{\to}12.8\%$ NBA, $73.3\%{\to}12.2\%$ Stock), while the performance gap---$0.017$ on Legal, $0.194$ on NBA, $0.111$ on Stock---quantifies how much of Temporal RAG's apparent skill was contamination rather than genuine forecasting. The gap is negligible on Legal, where leaked knowledge adds little, and largest on Stock, where nearly all of Temporal RAG's ranking ability traces to post-cutoff information. The performance change under enforcement is itself a per-task contamination diagnostic: it tells the practitioner how much of the reported performance was genuine.

%% file: sections/emnlp_related_work.tex

\section{Related Work}
\label{sec:related_work}

\paragraph{Data Contamination and Temporal Knowledge in LLMs.}
Data contamination---the overlap between training corpora and evaluation benchmarks---inflates performance and undermines reliability~\citep{brown2020language,carlini2021extracting,jiang2024contamination}. A related but distinct phenomenon is \emph{temporal} contamination: LLMs exhibit temporal blind spots~\citep{wallat2024temporal}, limited awareness of their own knowledge boundaries~\citep{pkezik2025llmlagbench,chenghaozhu2025your}, and degraded performance on temporally shifted inputs~\citep{agarwal2022temporal,dhingra2022time}. Recent work formalizes lookahead bias as a structural property of pretrained representations~\citep{sarkar2025lookahead} and proposes contamination-free evaluation through dynamic question generation~\citep{forecastbench2025,iclr2026forecasting}. Our framework addresses a complementary gap: identifying \emph{which} claims constitute leakage and quantifying \emph{how much} each influences the prediction via Shapley attribution.

\paragraph{LLM Backtesting and Temporal Mitigation.}
Backtesting LLMs on historical events risks conflating hindsight bias with genuine predictive signal~\citep{bailey2014determining,tetlock2014forecasting}, and studies in finance confirm that apparent forecasting gains often reflect a ``profit mirage'' driven by temporal leakage~\citep{li2025profit,yu2025livetradebench,chen2024does}. Training-time approaches address this at the model level: Time Machine GPT~\citep{drinkall2024timemachine}, ChronoGPT~\citep{he2025chronogpt}, and DATEDGPT~\citep{yan2026datedgpt} train on temporally partitioned corpora to eliminate leakage by construction, but require retraining for each cutoff and are limited to models the practitioner controls. At inference time, retrieval-augmented generation~\citep{lewis2020retrieval} is widely adopted but insufficient: date-filtered retrieval can amplify leakage by $2.5$--$4.7\times$ through unreliable metadata~\citep{ellahib2026datefilter}. Our framework is complementary: it operates on any LLM without retraining, combining claim-level verification with temporally-filtered retrieval to suppress both parametric and retrieval-borne contamination.

\paragraph{Shapley Attribution and Interpretable Evaluation.}
Shapley values~\citep{shapley1953value} provide axiomatically justified attribution. SHAP~\citep{lundberg2017unified} and its variants---KernelSHAP~\citep{covert2021improving} via weighted regression, Leverage SHAP~\citep{musco2025leverage} via leverage-score sampling with provable $(1+\epsilon)$ guarantees---make estimation practical; we adopt Leverage SHAP for claim-level attribution at scale. On the evaluation side, rationale faithfulness~\citep{deyoung2020eraser}, atomic-fact precision~\citep{min2023factscore}, search-augmented factuality evaluation~\citep{wei2024long}, and claim-level entailment verification~\citep{kamoi2023wice,thorne2018fact} provide complementary lenses on output quality. Our framework unifies attribution and verification with a temporal dimension: claims are decomposed, temporally classified, verified through date-aware search, and weighted by Shapley-derived decision impact.

%% file: sections/emnlp_conclusion.tex

\section{Conclusion}
\label{sec:conclusion}

We presented two contributions for reliable LLM backtesting. First, a model-agnostic evaluation framework decomposes prediction rationales into atomic claims, classifies their temporal provenance, and applies Shapley attribution to quantify each leaked claim's influence on the final decision, yielding Shapley-DCLR as an interpretable contamination metric. Second, TimeSPEC prevents leakage at inference time by coupling temporally-filtered retrieval with claim-level supervision under a closed-world constraint. A large-scale ablation across three LLMs and 2{,}769 forecasting instances confirms that retrieval and supervision are jointly necessary, and a three-task probe demonstrates that the performance cost of temporal enforcement scales with each task's dependence on post-cutoff information---diagnosing how much of the unfiltered baseline's skill was contamination.

%% file: sections/emnlp_appendix.tex

\newpage

\section{Dataset Details}
\label{app:datasets}

This section describes the datasets used in both experiments. Section~\ref{app:data_ablation} covers the large-scale ablation and Section~\ref{app:data_probe} covers the task-sensitivity probe.

\subsection{Large-Scale Ablation}
\label{app:data_ablation}

The ablation uses 2{,}769 binary forecasting questions partitioned into Treatment and Control groups by temporal accessibility relative to model knowledge cutoffs.

\paragraph{Treatment group.}
The Treatment group comprises 1{,}942 binary questions drawn from Autocast~\citep{zou2022forecasting}, a forecasting benchmark of 6{,}707 questions collected from three public forecasting tournaments (Metaculus, Good Judgment Open, and CSET Foretell). Questions span geopolitics, economics, science, and public health, each with a ground-truth resolution and a known resolution date. We retain the binary-outcome subset whose resolution dates precede the knowledge cutoff of every model tested (Qwen3.5-35B, Kimi-K2.5, Claude Sonnet~4), so parametric leakage from training data is possible.

\paragraph{Control group.}
The Control group comprises 827 binary questions drawn from ForecastBench~\citep{forecastbench2025}, a dynamic benchmark that ingests questions nightly from nine sources---four forecasting platforms (Manifold, Metaculus, Polymarket, RAND Forecasting Initiative) and five real-world datasets (ACLED, DBnomics, FRED, Wikipedia, Yahoo Finance)---and samples fresh question sets every two weeks. By construction, all Control questions resolve after every model's knowledge cutoff, so parametric leakage is impossible. This group serves three purposes: it validates that Shapley-DCLR does not produce false positives on leakage-free instances, isolates retrieval as a leakage channel by removing the parametric confound, and establishes a lower bound on achievable DCLR.

Table~\ref{tab:ablation_data} summarizes the two groups.

\begin{table*}[t]
\centering
\caption{Large-scale ablation dataset summary. Treatment questions admit parametric leakage; Control questions do not.}
\label{tab:ablation_data}
\footnotesize
\setlength{\tabcolsep}{5pt}
\begin{tabular}{@{}llccl@{}}
\toprule
\textbf{Group} & \textbf{Source} & \textbf{$N$} & \textbf{Format} & \textbf{Leakage Status} \\
\midrule
Treatment & Autocast~\citep{zou2022forecasting} & 1{,}942 & Binary & Possible (pre-cutoff resolution) \\
Control   & ForecastBench~\citep{forecastbench2025} & 827    & Binary & Impossible (post-cutoff resolution) \\
\midrule
Total     &              & 2{,}769 &        & \\
\bottomrule
\end{tabular}
\end{table*}

\subsection{Task-Sensitivity Probe}
\label{app:data_probe}

The probe comprises three forecasting tasks that span the leakage-sensitivity spectrum.

\paragraph{Stock return ranking.}
We construct 100 ranking instances from the COVID-19 market period (December~2019--June~2020). Each instance contains 5 S\&P~500 stocks from a single GICS sector; the task is to rank them by realized six-month return. The reference date is fixed at $t_{\text{ref}} = \text{December 1, 2019}$, which precedes public awareness of COVID-19 (China alerted the WHO on December~31, 2019). This design makes the task deliberately leakage-sensitive: accurate ranking requires knowledge of pandemic-driven market dynamics that were unknowable at $t_{\text{ref}}$. Performance is $(\rho+1)/2$, where $\rho$ is Spearman rank correlation~\citep{spearman1904proof}, normalized to $[0,1]$. Stock price data are from Yahoo Finance.\footnote{\url{https://finance.yahoo.com}}

\paragraph{NBA salary prediction.}
We compile 137 notable NBA free-agent contracts spanning 2019--2025 from public transaction records.\footnote{Salary data from Spotrac: \url{https://www.spotrac.com/nba}} The task is to predict each player's annual average value (AAV) in USD. Input features include the player's previous team, position, and career statistics. For each instance, $t_{\text{ref}}$ is set before the contract announcement, ensuring that the signed contract details---and contemporaneous peer deals---are post-cutoff. This task has intermediate leakage sensitivity: pre-cutoff evidence (player statistics, age, prior extensions, salary cap projections) is informative, but the exact contract values that an unfiltered model can memorize are post-cutoff. Performance is $1 - |\hat{y}-y|/|y|$, higher is better.

\paragraph{Legal case outcome prediction.}
We curate 100 U.S.\ Supreme Court cases from recent terms.\footnote{Case data from the Supreme Court Database and Oyez: \url{https://www.oyez.org}} The task is binary classification: predict whether the petitioner prevails. Input features include case name, parties, legal question, and factual background. For each case, $t_{\text{ref}}$ is set between oral argument and the decision date, so the ruling itself is post-cutoff while all briefing, argument transcripts, and prior precedent are available. This task is the least leakage-sensitive: judicial outcomes rely primarily on precedent established years or decades earlier, statutory text fixed at enactment, and judicial philosophies observable through prior opinions. Performance is $1 - (\hat{p}-y)^2$, where $\hat{p} \in [0,1]$ is the predicted probability and $y \in \{0,1\}$ the outcome, higher is better.

Table~\ref{tab:probe_data} summarizes the three tasks.

\begin{table*}[t]
\centering
\caption{Task-sensitivity probe dataset summary. Tasks are ordered from highest to lowest leakage sensitivity.}
\label{tab:probe_data}
\footnotesize
\setlength{\tabcolsep}{4pt}
\begin{tabular}{@{}llccll@{}}
\toprule
\textbf{Task} & \textbf{Type} & \textbf{$N$} & \textbf{Period} & \textbf{Cutoff Design} & \textbf{Metric} \\
\midrule
Stock   & Ranking        & 100 & Dec 2019--Jun 2020 & Period start      & $(\rho{+}1)/2$ ($\uparrow$) \\
Salary  & Regression     & 137 & 2019--2025         & Pre-signing       & $1{-}|\hat{y}{-}y|/|y|$ ($\uparrow$) \\
Legal   & Classification & 100 & Recent terms       & Pre-decision      & $1{-}(\hat{p}{-}y)^2$ ($\uparrow$) \\
\bottomrule
\end{tabular}
\end{table*}


\section{Claim Taxonomy: Full Specification}
\label{app:taxonomy_full}

Evaluating the leakage indicator $\ell(c; t_{\text{ref}}) = \mathbf{1}[\tau(c) > t_{\text{ref}}]$ for an arbitrary natural-language claim requires establishing the earliest date at which $c$ is publicly knowable, then comparing against $t_{\text{ref}}$. Most claims, however, fall into types whose leakage status is determined by structure alone, bypassing external lookup entirely. We formalize this observation as a nine-class taxonomy $\{\text{C0}, \ldots, \text{C8}\}$ that partitions all admissible factual claims. Table~\ref{tab:claim_taxonomy} lists all nine categories with examples drawn from our forecasting tasks and the verification rule applied by Phase~3 of the evaluation pipeline.

\begin{table*}[t]
\centering
\caption{Full nine-class claim taxonomy. The \emph{Status} column indicates how $\ell(c)$ is determined: deterministic categories (C0--C1, C7--C8) bypass external search entirely, while C2--C6 require date verification via $\tau(c)$. Horizontal rules separate the three operational tiers.}
\label{tab:claim_taxonomy}
\footnotesize
\setlength{\tabcolsep}{3pt}
\renewcommand{\arraystretch}{1.25}
\begin{tabular}{@{}c l p{5.5cm} l p{3.2cm}@{}}
\toprule
\textbf{ID} & \textbf{Type} & \textbf{Example Claim} & \textbf{Status} & \textbf{$\tau(c)$ Rule} \\
\midrule
C0 & Timeless Definition &
\textit{``The Brier score is $(p-o)^2$ for predicted probability $p$ and binary outcome $o$.''} &
Never leaked &
$\tau(c) = -\infty$; time-invariant by definition. \\[3pt]

C1 & Stable Entity Attribute &
\textit{``The U.S.\ Supreme Court is composed of nine justices.''} &
Usually safe &
Deterministic; attribute must be stable through $t_{\text{ref}}$. \\

\midrule

C2 & Time-Indexed State &
\textit{``As of Q3 2019, U.S.\ crude-oil demand had been declining for three consecutive quarters.''} &
Verify date &
Earliest source date corroborating the stated condition. \\[3pt]

C3 & Quantitative Measurement &
\textit{``Apple's FY2019 revenue was \$260.2B.''} &
Verify date &
Date of public release (e.g., 10-K filing date). \\[3pt]

C4 & Comparative / Relational &
\textit{``In 2019, Boeing's debt-to-equity ratio exceeded that of Airbus by $>$3$\times$.''} &
Verify date &
Latest source date among the comparands. \\[3pt]

C5 & Discrete Event &
\textit{``On Oct.~25, 2019, Microsoft won the \$10B JEDI cloud contract.''} &
Verify date &
Event date, or announcement date if unreported. \\[3pt]

C6 & Publication / Source &
\textit{``A Sep.~15, 2019 Reuters article reported Ford's market share at 13.4\%.''} &
Verify date &
Publication date of the cited source. \\

\midrule

C7 & Target Outcome &
\textit{``Durant signed a 4-yr, \$164M deal with the Nets.''}\; (when salary AAV is the prediction target) &
Always leaked &
$\tau(c) = t_E > t_{\text{ref}}$ by task definition. \\[3pt]

C8 & Post-Outcome Consequence &
\textit{``The signing pushed the Nets' 2019--20 payroll above the luxury-tax threshold.''} &
Always leaked &
$\tau(c) > t_E > t_{\text{ref}}$; exists only after the target resolves. \\

\bottomrule
\end{tabular}
\end{table*}

The taxonomy satisfies two properties that together guarantee a well-defined classification for every claim.

\begin{proposition}[Exclusivity]
\label{prop:exclusivity}
Let $\mathcal{C} = \{\text{C0}, \ldots, \text{C8}\}$ be the nine claim categories, and define the strict precedence order
\begin{align}
\text{C7} \succ \text{C8} \succ \text{C6} \succ \text{C3} \succ \text{C4} \notag\\
{}\succ \text{C5} \succ \text{C2} \succ \text{C1} \succ \text{C0}.
\label{eq:precedence}
\end{align}
For any factual claim $c$ that satisfies the membership criteria of two or more categories $\kappa_1, \kappa_2 \in \mathcal{C}$, the classifier assigns the unique label $\kappa^* = \arg\max_{\succ}\{\kappa_1, \kappa_2\}$. Thus every claim receives exactly one category.
\end{proposition}

\begin{justification}
The precedence order ranks categories by leakage severity: C7--C8 (deterministically leaked) dominate C2--C6 (date-dependent), which in turn dominate C0--C1 (deterministically safe). Within each tier, higher precedence goes to categories carrying stronger temporal signal---for instance, the target outcome (C7) outranks post-outcome consequences (C8), and publication-sourced claims (C6) outrank generic time-indexed states (C2). When a claim matches multiple categories, the $\arg\max$ under $\succ$ selects the unique highest-severity label, so no ambiguity arises.
\end{justification}

\begin{proposition}[Completeness]
\label{prop:completeness}
For any admissible factual claim $c$ extracted from a prediction rationale, there exists at least one $\kappa \in \mathcal{C}$ whose membership criteria $c$ satisfies. Completeness is guaranteed by C2 (Time-Indexed State), which serves as the residual class: any time-dependent factual assertion not captured by C3--C8 falls into C2 by default.
\end{proposition}

\begin{justification}
Categories C3--C8 capture specific claim types (quantitative measurements, events, publications, target outcomes, and their consequences), while C0--C1 capture time-invariant facts. Any remaining factual claim with temporal content---a condition, status, or trend indexed to a specific time period---satisfies C2's membership criterion by definition. Because C2 subsumes all temporal assertions not already captured by the specialized categories, no admissible claim is left unclassified.
\end{justification}

Together, Propositions~\ref{prop:exclusivity} and~\ref{prop:completeness} ensure that the taxonomy induces a partition over all admissible claims. This partition yields three \emph{operational tiers} that directly govern the verification cost within our leakage detection framework. Categories C0 and C1 are \emph{deterministically safe}: their temporal grounding satisfies $\tau(c) \leq t_{\text{ref}}$ by definition, so the pipeline assigns $\ell(c) = 0$ without invoking any external search. Symmetrically, C7 and C8 are \emph{deterministically leaked}: the target event and its downstream effects are post-cutoff by construction, so $\ell(c) = 1$ is assigned immediately. Only claims in the intermediate categories C2--C6 require date verification---the pipeline retrieves the earliest publicly available source for each claim and compares its publication date against $t_{\text{ref}}$ to determine $\ell(c)$. Empirically, the two deterministic tiers account for 30--50\% of extracted claims across our experiments, yielding a corresponding reduction in the number of search queries required per instance.

\paragraph{Worked example.}
We illustrate the taxonomy on a real instance from our large-scale ablation. The forecasting question \textit{``Will Elon Musk be sanctioned for Tesla buyout tweet?''} has reference date $t_{\text{ref}} = \text{September~1, 2018}$. Qwen3.5-35B produces a rationale containing seven atomic claims. Three claims describe pre-cutoff events---Musk's August~7 tweet, the subsequent stock-price surge, and the SEC's August~22 investigation announcement---all classified \textbf{C0} and assigned $\ell(c)=0$ deterministically without search. Three further claims reference the SEC settlement: the model declares dates in late August, but date verification corrects these to September~29 (post-cutoff), classifying all three as \textbf{C4} with $\ell(c) = 1$. The remaining claim---\textit{``The settlement required Musk to pay a \$20M fine, which exceeds the \$1M threshold specified in the question''}---directly reveals the prediction target and is classified \textbf{C7} deterministically. This claim also qualifies as C3 (quantitative measurement), but the precedence order (C7 $\succ$ C3, Eq.~\ref{eq:precedence}) assigns C7 as the unique label.

The resulting Shapley-DCLR is $0.462$: nearly half of the decision-driving reasoning is contaminated, with the C7 claim carrying the highest Shapley value ($\phi = 0.232$). Operationally, only the three C4 claims require external search; the four deterministic claims (three C0 and one C7) are resolved by category alone, reducing the verification budget by 57\%.

\section{Method Implementations}
\label{app:implementations}

This section details the evaluation pipeline (Section~\ref{app:eval_impl}) and the four baseline generation methods (Section~\ref{app:baseline_impl}). Code and configurations are publicly available.\footnote{\url{https://anonymous.4open.science/r/TimeSPEC-6E24}} Prompt specifications are in Appendix~\ref{app:prompts}.

\subsection{Evaluation Pipeline}
\label{app:eval_impl}

The evaluation pipeline runs post-hoc on the rationale produced by any generation method. It takes as input a structured rationale containing atomic evidence claims (each with a \texttt{fact}, \texttt{source\_date}, and \texttt{id}) and the instance's reference date $t_{\text{ref}}$, and outputs per-claim leakage decisions and Shapley-weighted decision-critical leakage rates.

\paragraph{Claim Extraction and Taxonomy Classification (Phase~1).}
We employ Claude Sonnet~4\footnote{\url{https://openrouter.ai/anthropic/claude-sonnet-4}} ($T{=}0$, \texttt{max\_tokens}${=}8000$) to decompose each rationale into atomic claims. Each extraction call receives the rationale text, task description, event description, and $t_{\text{ref}}$ as context. The extractor outputs structured objects containing claim text, original text span, temporal reference, C0--C8 category label, and category reasoning. The taxonomy classification follows Appendix~\ref{app:taxonomy_full} and applies the precedence order (Eq.~\ref{eq:precedence}) when a claim matches multiple categories.

\paragraph{Shapley Value Computation (Phase~2).}
Shapley values are computed via Leverage SHAP~\citep{musco2025leverage} following the reference implementation.\footnote{\url{https://github.com/rtealwitter/leverageshap}} The coalition sampler uses uniform per-size weights with paired complements and without-replacement sampling. The budget parameter is $C{=}15$ with a cap of 1{,}200 coalitions; for instances with $n \leq 8$ claims, exact enumeration ($2^n$ coalitions) is used instead. Each coalition is evaluated with a closed-world prompt that provides only the claims in the coalition subset; the prediction model matches the original generator (same OpenRouter model ID, $T{=}0$). The projected weighted regression enforces the efficiency axiom ($\sum_i \phi_i = v(N) - v(\emptyset)$) by construction. Coalition predictions are cached to avoid redundant model calls when multiple coalitions share the same claim subset. Concurrency is set to 8 parallel coalition evaluations.

\paragraph{Leakage Detection (Phase~3).}
Leakage detection applies the three-tier routing from Appendix~\ref{app:taxonomy_full} to minimize verification cost. C7 and C8 claims are marked leaked deterministically; C0 and C1 claims are marked safe deterministically. Only C2--C6 claims undergo date verification. For these, a taxonomy-aware LLM verifier (Gemini~2.0 Flash Lite,\footnote{\url{https://openrouter.ai/google/gemini-2.0-flash-lite-001}} $T{=}0$) first assesses whether the declared source date is plausible given the claim content. Claims with implausible or missing dates are escalated to search-based verification via Perplexity API\footnote{\url{https://www.perplexity.ai}}: the pipeline retrieves up to 3 results per claim and a second LLM pass extracts the earliest reliable public availability date from the search snippets. Coarse temporal references are normalized conservatively (e.g., ``2023'' $\to$ 2023-12-31, ``Q3 2023'' $\to$ 2023-09-30) to avoid false negatives. The final leakage decision compares the effective source date against $t_{\text{ref}}$.

\subsection{Generation Methods}
\label{app:baseline_impl}

The four methods occupy the cells of a $2{\times}2$ (Search, Supervision) grid, as described in Section~\ref{sec:experiments}. All methods share a common generation backbone accessed through OpenRouter,\footnote{\url{https://openrouter.ai}} with $T{=}0$ and \texttt{max\_tokens}${=}4096$. The pipeline exploits two shared generation phases to minimize redundant API calls: Phase~1A generates rationales without retrieval context (shared by Temporal Hint and Self-Correction), and Phase~1B generates rationales with retrieval context (shared by Temporal RAG and TimeSPEC). Output parsing and schema validation are identical across all methods.

\paragraph{Temporal Hint ($-$Search, $-$Supervision).}
The simplest baseline: the model receives the task input and a prompt-level temporal constraint (``Your knowledge cutoff is \texttt{\{date\}}. Only use information available on or before this date.'') and produces a single-pass prediction with rationale. No search tools, no supervision, no regeneration. This tests whether prompt engineering alone can prevent temporal leakage from the model's parametric memory.

\paragraph{Self-Correction ($-$Search, $+$Supervision).}
Self-Correction extends Temporal Hint with an LLM-only claim-level supervisor. After initial generation (shared Phase~1A), the supervisor classifies each evidence claim under the C0--C8 taxonomy and flags leaked claims using the LLM verifier alone (no external search). If leaked claims are detected, the method triggers regeneration: the model receives the validated (non-leaked) claims from all prior cycles and generates a new rationale. The cycle repeats up to \texttt{max\_regen\_cycles}${=}1$ time. When any leakage was detected across cycles, a closed-world aggregator synthesizes the final prediction from only the union of validated claims (deduplicated by fact text), ensuring no leaked content persists. The key architectural difference from TimeSPEC is the absence of retrieval: the supervisor operates entirely on the model's own claims without external date verification.

\paragraph{Temporal RAG ($+$Search, $-$Supervision).}
Temporal RAG augments the generation prompt with date-filtered retrieval context but applies no post-hoc claim verification. The retrieval pipeline first generates task-specific search queries via an LLM, then issues them to Perplexity API with a temporal cutoff constraint ($t < t_{\text{ref}}$), retrieves up to 5 results per query, and assembles a context block (budget: 3{,}000 characters) that is prepended to the generation prompt. The model produces a single-pass prediction conditioned on both the context and the temporal hint. This method tests whether retrieval with API-level date filtering suffices to prevent leakage---our experiments show it does not, as post-cutoff content passes through unreliable metadata~\citep{ellahib2026datefilter}.

\paragraph{TimeSPEC ($+$Search, $+$Supervision).}
The full pipeline described in Section~\ref{sec:timespec}. TimeSPEC combines the retrieval context of Temporal RAG with the claim-level supervision of Self-Correction, and adds two architectural enhancements. First, the supervisor uses \emph{search-escalated verification}: after the initial LLM-only taxonomy pass, claims with uncertain or borderline dates are escalated to Perplexity search for independent date confirmation, then re-verified against the search snippets by a second LLM pass. This catches cases where the model backdates post-cutoff events (as illustrated in the worked example of Appendix~\ref{app:taxonomy_full}). Second, a \emph{defense-in-depth evidence scrub} runs after the aggregator: any claim whose source date exceeds $t_{\text{ref}}$ or that was flagged by the last supervisor pass is stripped from the final evidence list, guarding against the aggregator hallucinating fresh post-cutoff facts. Regeneration follows the same cycle as Self-Correction (\texttt{max\_regen\_cycles}${=}1$), and the closed-world aggregator is always invoked when any leakage was detected.

Table~\ref{tab:method_comparison} summarizes the architectural differences across the four methods.

\begin{table*}[t]
\centering
\caption{Architectural comparison of the four generation methods. \cmark\ indicates the component is present; -- indicates absent.}
\label{tab:method_comparison}
\footnotesize
\setlength{\tabcolsep}{3pt}
\begin{tabular}{@{}lcccccc@{}}
\toprule
\textbf{Method} & \textbf{Prompt} & \textbf{RAG} & \textbf{Supervisor} & \textbf{Search} & \textbf{Regen} & \textbf{Aggregator} \\
\midrule
Temporal Hint   & \cmark & --     & --          & --         & --     & -- \\
Self-Correction & \cmark & --     & LLM-only    & --         & \cmark & \cmark \\
Temporal RAG    & \cmark & \cmark & --          & --         & --     & -- \\
TimeSPEC        & \cmark & \cmark & LLM+Search  & \cmark     & \cmark & \cmark \\
\bottomrule
\end{tabular}
\end{table*}


\section{Leakage Concentration in Decision-Critical Claims}
\label{app:topk_leakage}

Shapley-DCLR (Eq.~\ref{eq:shapley_dclr}) quantifies the aggregate fraction of decision weight attributable to leaked claims, but does not reveal whether contamination concentrates among the most influential claims or spreads thinly across many low-impact ones. The distinction matters: a 20\% DCLR carried entirely by the single highest-Shapley claim poses a greater integrity risk than the same 20\% distributed across peripheral reasoning. We examine this concentration effect by computing Top-$K$ leakage rates across all four baselines and three tasks from the task-sensitivity probe (Section~\ref{subsec:analysis}).

\paragraph{Definition.}
For each instance, we rank claims by Shapley value magnitude $|\phi_i|$ in descending order and compute the leakage rate among the top-$K$ most influential claims:
\begin{equation}
\text{Top-}K\text{ Leakage} = \frac{1}{K} \sum_{i=1}^{K} \ell\bigl(c_{(i)}\bigr),
\label{eq:topk}
\end{equation}
where $c_{(i)}$ denotes the claim with the $i$-th highest Shapley magnitude and $\ell(\cdot) \in \{0,1\}$ is the leakage indicator. Instances with zero leaked claims receive Top-$K = 0$ for all $K$. We report the mean across all instances per task, alongside Shapley-DCLR for reference. If Top-$K$ exceeds DCLR, leaked claims are over-represented at the top of the influence ranking.

Table~\ref{tab:topk_leakage} reports the results. All runs use Qwen3.5-35B.

\begin{table*}[t]
\centering
\caption{Leakage concentration among the most decision-critical claims. \emph{DCLR} is the Shapley-weighted aggregate from Section~\ref{subsec:analysis}; \emph{Top-$K$} is the fraction of the $K$ highest-Shapley claims that are leaked (Eq.~\ref{eq:topk}), averaged across instances. All values in \%. Top-1 $>$ DCLR indicates that leakage concentrates at the most influential position. Bold marks the lowest value per column within each task group.}
\label{tab:topk_leakage}
\footnotesize
\setlength{\tabcolsep}{3pt}
\resizebox{\textwidth}{!}{%
\begin{tabular}{@{}l cccc cccc cccc@{}}
\toprule
& \multicolumn{4}{c}{\textbf{Legal} ($n{=}100$)} & \multicolumn{4}{c}{\textbf{NBA Salary} ($n{=}137$)} & \multicolumn{4}{c}{\textbf{Stock} ($n{=}100$)} \\
\cmidrule(lr){2-5} \cmidrule(lr){6-9} \cmidrule(lr){10-13}
\textbf{Method} & DCLR\,$\downarrow$ & Top-1\,$\downarrow$ & Top-3\,$\downarrow$ & Top-5\,$\downarrow$ & DCLR\,$\downarrow$ & Top-1\,$\downarrow$ & Top-3\,$\downarrow$ & Top-5\,$\downarrow$ & DCLR\,$\downarrow$ & Top-1\,$\downarrow$ & Top-3\,$\downarrow$ & Top-5\,$\downarrow$ \\
\midrule
Temporal Hint   & 11.7 & 17.0 & 11.3 & 10.1  & 58.5 & 70.8 & 50.4 & 48.2  & 26.0 & 28.0 & 26.7 & 26.2 \\
Self-Correction & \textbf{5.6} & 7.0 & \textbf{5.0} & \textbf{4.0}  & \textbf{11.3} & \textbf{11.7} & \textbf{9.7} & \textbf{9.0}  & \textbf{10.2} & \textbf{7.0} & 8.7 & 8.6 \\
Temporal RAG    & 32.5 & 41.0 & 35.3 & 30.7  & 61.7 & 75.2 & 59.1 & 52.0  & 73.3 & 80.0 & 74.3 & 72.8 \\
TimeSPEC        & 5.9 & \textbf{5.0} & 5.7 & 5.8  & 12.8 & 14.6 & 12.5 & 10.6  & 12.2 & \textbf{7.0} & \textbf{7.0} & \textbf{7.0} \\
\bottomrule
\end{tabular}}%
\end{table*}

\paragraph{Legal: low leakage, no concentration.}
All four methods produce DCLR below 12\%, and Top-$K$ rates remain in the same range. The gap between Top-1 and DCLR is modest even for Temporal RAG (41.0\% vs.\ 32.5\%), and the two supervised methods bring Top-1 below 7\%. Legal reasoning relies on precedent and statutory text available well before the decision date, so the few leaked claims that appear are not systematically the most influential ones. Supervision reduces both the aggregate and the peak with little differential effect.

\paragraph{NBA Salary: strong concentration in unsupervised methods.}
Salary prediction exhibits the most pronounced concentration effect. Temporal Hint's Top-1 leakage (70.8\%) exceeds its DCLR (58.5\%) by 12 percentage points: in over 70\% of instances, the single most influential claim is leaked---typically the actual signed contract value or a contemporaneous comparable deal that the model memorized from training data. Temporal RAG amplifies this further (Top-1 = 75.2\%, DCLR = 61.7\%), as retrieved post-cutoff contract details surface as high-Shapley evidence. Self-Correction reduces Top-1 to 11.7\% and TimeSPEC to 14.6\%, both close to their respective DCLRs (11.3\%, 12.8\%), indicating that claim-level supervision removes leaked claims specifically from the high-impact positions rather than merely suppressing peripheral contamination. The declining Top-1 $\to$ Top-5 gradient for Temporal Hint (70.8\% $\to$ 48.2\%) and Temporal RAG (75.2\% $\to$ 52.0\%) confirms that leakage concentrates at the top: diluting across more claims lowers the rate, meaning the very first claim is disproportionately likely to be contaminated.

\paragraph{Stock: retrieval creates extreme top-rank contamination.}
On the most leakage-sensitive task, Temporal RAG exhibits 80.0\% Top-1 leakage---in four out of five instances, the single most influential claim is leaked knowledge of COVID-era market movements. The rate barely declines to 72.8\% at Top-5, indicating that contamination saturates the entire upper stratum of the influence ranking, consistent with this method's 73.3\% DCLR. Temporal Hint shows a weaker but still present concentration effect (Top-1 = 28.0\% vs.\ DCLR = 26.0\%). Both supervised methods sharply reduce Top-1 to 7.0\%. TimeSPEC maintains a flat 7.0\% across all $K$ values, while Self-Correction shows slightly higher Top-3 and Top-5 (8.7\%, 8.6\%)---indicating that TimeSPEC's search-escalated verification removes contamination more thoroughly from the full upper stratum, not just the peak position.

\paragraph{Summary.}
Across all three tasks, unsupervised methods exhibit a consistent pattern: Top-1 leakage exceeds DCLR, confirming that leaked claims are over-represented at the most influential position. The concentration effect is strongest on tasks with high leakage sensitivity (Salary, Stock) and weakest where valid pre-cutoff evidence suffices (Legal). Claim-level supervision---whether LLM-only (Self-Correction) or search-escalated (TimeSPEC)---eliminates this concentration: Top-$K$ rates converge to DCLR, meaning residual leakage no longer preferentially occupies decision-critical positions. The practical implication is that DCLR alone understates the integrity risk of unsupervised predictions: the contaminated fraction disproportionately drives the final answer.

\section{Case Studies}
\label{app:case_study}

We present one representative instance from each task to illustrate how temporal leakage manifests in model rationales, how Shapley-DCLR exposes the contaminated reasoning that drives predictions, and how TimeSPEC suppresses it. All four methods run on Qwen3.5-35B.

\subsection{Stock Return Ranking: Healthcare Sector}

We examine a ranking instance from the Health Care sector: five stocks---Moderna (MRNA), Bio-Techne (TECH), Becton Dickinson (BDX), Stryker (SYK), and Dexcom (DXCM)---ranked by six-month return from December~1, 2019 to June~2, 2020. The reference date is $t_{\text{ref}} = \text{December 1, 2019}$, which precedes any public knowledge of COVID-19 (China alerted the WHO on December~31, 2019). The ground-truth ranking by realized returns is MRNA $>$ DXCM $>$ TECH $>$ SYK $>$ BDX, driven almost entirely by the pandemic: Moderna's mRNA vaccine program propelled MRNA to the top, while medical-device names lagged.

Temporal Hint and Temporal RAG both achieve a perfect ranking ($\rho = 1.0$), and Self-Correction achieves $\rho = 0.95$ (swapping only BDX and SYK). At face value, this suggests strong forecasting ability. Shapley-DCLR tells a different story. Temporal Hint's DCLR is 55.7\%: the rationale contains nine extracted claims, five of which are leaked. The most influential leaked claims state that ``Moderna's stock price rose from roughly \$100 in December~2019 to over \$200 by late February~2020'' (C8, $\phi = 0.124$), that ``on January~31, 2020, Moderna announced the successful isolation of the SARS-CoV-2 virus and the design of its mRNA vaccine candidate'' ($\phi = 0.126$), and that ``the global outbreak of COVID-19 in early~2020 created unprecedented demand for vaccine development'' ($\phi = 0.112$). These claims---Moderna's post-cutoff stock trajectory, the vaccine announcement two months after $t_{\text{ref}}$, and the pandemic itself---encode the very mechanism that produced the ground-truth ranking. The remaining safe claims describe pre-cutoff fundamentals for the non-biotechnology names (Bio-Techne's steady growth, Dexcom's CGM market position, BDX and SYK as lower-volatility device companies), which alone do not predict Moderna's pandemic-era dominance. The 55.7\% DCLR quantifies this: over half the decision weight rests on information that did not exist on December~1, 2019, and the ``perfect'' ranking is a product of temporal contamination rather than analytical skill.

Self-Correction reduces DCLR from 55.7\% to 30.5\% by removing the three most explicit post-cutoff claims---the COVID-19 outbreak, the Moderna stock trajectory, and the January~2020 vaccine announcement. Two subtler leaked claims survive: one referencing Moderna ``trading at approximately \$100 per share'' with a verified source date in January~2020, and another alluding to a ``vaccine development catalyst'' sourced from March~2020. Both are phrased as general market observations rather than dated events, evading the LLM-only verifier. The ranking remains near-perfect (0.95), indicating that even partial leakage suffices for accurate ranking when the leaked information encodes the dominant market driver.

Temporal RAG amplifies contamination to its extreme: DCLR = 100\%, with all eight extracted claims leaked. The retrieved content includes explicit stock prices from December~2019 through December~2020, pandemic-driven sector commentary, and year-end performance summaries. The highest-Shapley claim ($\phi = 0.193$) states that established companies' ``returns are expected to be lower than Moderna's explosive growth driven by COVID-19 vaccine development,'' directly encoding the pandemic narrative as retrieved evidence. The second-highest claim references Moderna's ``major surge in 2020'' ($\phi = 0.178$). API-level date filtering failed entirely: post-cutoff articles with unreliable metadata passed through, and the model assembled a rationale constructed entirely from future information. The perfect ranking carries zero predictive content.

TimeSPEC achieves DCLR = 0.0\%. Its search-escalated verification blocks all pandemic-era content, and the validated claims are grounded exclusively in pre-cutoff fundamentals: Moderna's post-IPO recovery in 2019, Bio-Techne's fiscal~2019 results released in July~2019, Stryker's February~2019 10-K filing, Dexcom's November~2019 market capitalization, and BDX's 2019 corporate governance activity. The resulting ranking places MRNA first---justified by its strong 2019 recovery trajectory---but ranks DXCM fourth rather than second, because no pre-cutoff evidence predicts the pandemic-driven surge in continuous glucose monitoring demand. The lower Spearman correlation ($\rho = 0.85$, performance = 0.85) reflects what an analyst could legitimately infer from Q3~2019 earnings alone. This is precisely the outcome honest backtesting should produce: without foreknowledge of a black swan event, perfect ranking of pandemic beneficiaries is unjustified, and the ``underperformance'' is a correct diagnostic that the task's apparent predictability was an artifact of contamination.

\subsection{NBA Salary Prediction: Kevin Durant (2019)}

Kevin Durant (SF, Golden State Warriors) entered the 2019 free agency and signed a 4-year, \$164.3M contract with the Brooklyn Nets on June~30, 2019 (AAV: \$41.06M). The reference date is $t_{\text{ref}} = \text{June 23, 2019}$---one week before the signing. This narrow temporal margin makes the case particularly revealing: pre-cutoff information (CBA rules, Durant's service years, market projections) is sufficient for a reasonable estimate, but the exact contract terms were announced only seven days after $t_{\text{ref}}$.

Temporal Hint, Self-Correction, and Temporal RAG all predict within 0.2\% of the ground truth (\$41.0M, \$41.0M, and \$41.08M respectively), while TimeSPEC predicts \$29.5M (28.2\% relative error). A naive comparison would conclude that the first three methods vastly outperform TimeSPEC. Shapley-DCLR reveals why this conclusion is misleading.

Temporal Hint's rationale contains five claims, four of which are temporally valid: Durant's free-agent status, the Nets' pursuit, the CBA maximum salary for a 10$+$-year veteran (approximately \$39.9M), and Durant's 11~years of service. The single leaked claim (C7, Shapley rank~1, $\phi = 0.222$) states that ``reports in late June~2019 indicated that Kevin Durant agreed to a 4-year, \$164~million contract with the Brooklyn Nets''---the target outcome itself, with a verified source date of June~30, seven days after $t_{\text{ref}}$. This claim carries the highest Shapley value among all five claims, indicating that the model's near-perfect prediction is anchored by memorized knowledge of the actual signing. DCLR = 22.2\%: over one-fifth of the decision weight traces to this single leaked fact.

Self-Correction produces identical results (DCLR = 22.3\%) because the LLM-only supervisor fails to catch the leaked claim. The model backdated the source to ``late June~2019'' without specifying a precise date, and without external search the supervisor cannot distinguish pre-cutoff rumors from the post-cutoff announcement. This illustrates a limitation of LLM-only verification: when the temporal margin is narrow---seven days in this case---the supervisor lacks the independent evidence needed to detect subtle backdating.

Temporal RAG doubles the contamination. Its retrieval pipeline surfaces two explicitly post-cutoff claims: ``on June~30, 2019, Kevin Durant announced that he planned to sign with the Brooklyn Nets'' ($\phi = 0.111$) and ``Durant signed with Brooklyn on July~7 on a four-year, \$164.3~million contract, in a sign-and-trade deal'' (Shapley rank~1, $\phi = 0.348$). The latter claim carries the highest decision impact by a wide margin---it effectively provides the answer as retrieved evidence, explaining the near-zero prediction error. DCLR rises to 45.8\%. The two safe claims (CBA salary rules and Durant's championship credentials) carry lower Shapley weight, confirming that the prediction is driven primarily by leaked contract details rather than market-rate inference.

TimeSPEC's search-escalated verification correctly identifies and blocks all post-cutoff contract details. The validated evidence grounds the prediction in CBA maximum-salary rules and Durant's service-year eligibility, yielding \$29.5M---the approximate CBA maximum for a 10$+$-year veteran without supermax eligibility. The 28.2\% gap between this estimate and the actual \$41.06M AAV reflects the difference between what CBA rules predict and the above-maximum deal Durant negotiated. The key point is that TimeSPEC's ``error'' is honest: the prediction is fully grounded in pre-cutoff evidence (DCLR = 0.0\%), while the three other methods' accuracy is partially explained by access to the actual contract value. Shapley-DCLR makes this distinction interpretable: it identifies the specific leaked claim (the contract announcement) and quantifies its contribution to the prediction (22--35\% of decision weight depending on the method), rather than merely flagging the instance as contaminated.

\subsection{Legal Case Prediction: Hargress v.\ SSA (2018)}

Joyce Hargress appealed the denial of her Social Security disability benefits to the U.S.\ Court of Appeals for the Eleventh Circuit. The ALJ had found that Hargress, who alleged disability due to type~II diabetes, tiredness, and anxiety, could perform sedentary unskilled work. The Appeals Council denied review. The reference date is $t_{\text{ref}} = \text{December 29, 2017}$; the court affirmed the denial on February~27, 2018, in a unanimous opinion (respondent prevails, $y = 0$).

Temporal Hint and Temporal RAG both predict $\hat{p} = 0.00$ (perfect accuracy), while Self-Correction and TimeSPEC both predict $\hat{p} = 0.15$ (slightly lower but still high accuracy). Shapley-DCLR again inverts the apparent ranking.

Temporal Hint achieves DCLR = 100\%: every one of its five extracted claims describes the court's decision. The first claim (C6) states that ``the case was decided by the Eleventh Circuit on January~11, 2018,'' while the remaining four (all C8) describe the court affirming the denial, finding substantial evidence to support the ALJ, and rejecting Hargress's arguments. Each effective source date falls in January~2018---two weeks after $t_{\text{ref}}$. The five Shapley values are uniform ($\phi = 0.200$ each), and since every claim is leaked, DCLR = 100\%. The model's perfect prediction ($\hat{p} = 0.00$) is not forecasting---it is recall of the known outcome. Shapley-DCLR correctly identifies that the entire decision-driving reasoning chain is contaminated.

Self-Correction detects and removes all five leaked claims. The supervisor correctly classifies them as post-cutoff (C6 and C8 categories are deterministically leaked under the taxonomy), triggers regeneration, and the model produces five replacement claims---all temporally valid. Two C0 claims describe the standard of review for Social Security disability appeals and the Eleventh Circuit's historically deferential posture toward ALJ findings. Two C1 claims summarize the ALJ's residual functional capacity determination and the Appeals Council's denial of review. All source dates fall between 2015 and 2017, well before $t_{\text{ref}}$. The prediction shifts from $\hat{p} = 0.00$ to $\hat{p} = 0.15$: the model assigns a low but nonzero probability to the petitioner, reflecting legitimate analytical uncertainty about whether the ALJ's credibility determination survives appellate review. DCLR = 0.0\%.

Temporal RAG reintroduces contamination. Its retrieval pipeline surfaces court records containing the outcome: three of four claims are leaked (DCLR = 78.8\%), including the highest-Shapley claim ($\phi = 0.288$), which states that ``the Eleventh Circuit affirmed the denial'' with a January~2018 source date. The single safe claim describes the ALJ's 2014 residual functional capacity finding. The model predicts $\hat{p} = 0.00$ with certainty derived from the retrieved ruling.

TimeSPEC blocks all post-cutoff court records through search-escalated verification. The validated evidence consists of the procedural record available before the appellate decision, and the prediction ($\hat{p} = 0.15$) matches Self-Correction's output at DCLR = 0.0\%.

This case illustrates an important edge case for the legal task. On average, legal prediction exhibits the lowest DCLR across all methods (Section~\ref{subsec:analysis}), because most instances are predicted from precedent and procedural history established years before the decision. This specific instance, however, shows that individual legal cases can exhibit extreme contamination when the model directly recalls the ruling. The distinction between $\hat{p} = 0.00$ (contaminated certainty from recalled outcome) and $\hat{p} = 0.15$ (legitimate uncertainty from pre-cutoff procedural analysis) is precisely what Shapley-DCLR is designed to detect: not merely whether leakage is present, but whether it drives the prediction. In this case, it drives the entire prediction---and the evaluation framework correctly reports DCLR = 100\%.


\onecolumn

\section{Prompt Specifications}
\label{app:prompts}

All prompts are reproduced from the implementation source code for reproducibility. Template variables are shown as \texttt{\{variable\}}. Section~\ref{app:prompt_gen} covers generation prompts shared across the four methods; Section~\ref{app:prompt_sup} covers supervision and regeneration prompts used by Self-Correction and TimeSPEC; Section~\ref{app:prompt_eval} covers evaluation pipeline prompts. Table~\ref{tab:prompt_usage} summarizes which prompts each method uses.

\begin{table}[h]
\centering
\caption{Prompt usage across methods. \cmark\ = used; -- = not used.}
\label{tab:prompt_usage}
\footnotesize
\setlength{\tabcolsep}{3pt}
\begin{tabular}{@{}lcccc@{}}
\toprule
\textbf{Prompt} & \textbf{T.~Hint} & \textbf{Self-Corr.} & \textbf{T.~RAG} & \textbf{TimeSPEC} \\
\midrule
Generation (System + User) & \cmark & \cmark & \cmark & \cmark \\
RAG Context Block          & --     & --     & \cmark & \cmark \\
Query Generation           & --     & --     & \cmark & \cmark \\
Taxonomy Verification      & --     & \cmark & --     & \cmark \\
Search Re-Verification     & --     & --     & --     & \cmark \\
Regeneration               & --     & \cmark & --     & \cmark \\
Aggregation                & --     & \cmark & --     & \cmark \\
\bottomrule
\end{tabular}
\end{table}

\makeatletter
\renewcommand{\_}{\textunderscore\discretionary{}{}{}}
\makeatother

\subsection{Generation Prompts}
\label{app:prompt_gen}

All four methods share the same generation backbone. The system prompt embeds a strict temporal constraint and a three-step evidence--reasoning--prediction procedure. The user prompt supplies the question and optional RAG context. Task-specific output schemas are appended to the system prompt.

\paragraph{Shared System Prompt.}
Used by all four methods (B1--B4). Temporal RAG and TimeSPEC receive additional RAG context in the user message; the system prompt is identical.

\begin{tcolorbox}[colback=red!3, colframe=red!50!black, fonttitle=\bfseries\small, title=Generation System Prompt]
\small
You are an expert forecaster. Your task is to answer a prediction question using only information that was publicly available before the specified knowledge cutoff date.\textbackslash n\textbackslash n
STRICT TEMPORAL CONSTRAINT:\textbackslash n
You may ONLY use information that was publicly available BEFORE \texttt{\{cutoff\_date\}}. Do NOT use knowledge of events, announcements, or developments from on or after \texttt{\{cutoff\_date\}}. If uncertain whether a piece of information predates the cutoff, omit it.\textbackslash n\textbackslash n
EVIDENCE INTEGRITY:\textbackslash n
Evidence must be factually genuine. Do not fabricate numbers, dates, or events. Each evidence item must reflect a real, publicly known fact. If you are uncertain about a specific value, state the fact qualitatively rather than inventing a precise figure.\textbackslash n\textbackslash n
You MUST follow this three-step procedure:\textbackslash n\textbackslash n
STEP 1 --- EVIDENCE:\textbackslash n
List every verifiable fact you will use. Requirements:\textbackslash n
- Each fact must be concrete and specific --- not an opinion or prediction.\textbackslash n
- Each fact must have been publicly known before \texttt{\{cutoff\_date\}}.\textbackslash n
- For EACH fact, include a source\_date field: the YYYY-MM-DD date when this information was FIRST PUBLICLY ANNOUNCED.\textbackslash n\textbackslash n
STEP 2 --- REASONING:\textbackslash n
Write a reasoning paragraph that cites evidence by number (e.g.\ [1], [3]).\textbackslash n
- You MUST cite every evidence item at least once.\textbackslash n
- Do NOT introduce any new facts not listed in your evidence.\textbackslash n\textbackslash n
STEP 3 --- PREDICTION:\textbackslash n
Give your final prediction based solely on the reasoning above.\textbackslash n
CRITICAL: You MUST always output the JSON object below, even under uncertainty.\textbackslash n
\texttt{\{output\_spec\}}
\end{tcolorbox}

\paragraph{User Message Structure.}
The user message is constructed programmatically. For Temporal Hint and Self-Correction, it contains only the question and background. For Temporal RAG and TimeSPEC, a \texttt{RETRIEVED CONTEXT (all pre-cutoff)} block is prepended with search results assembled by the RAG pipeline.

\begin{tcolorbox}[colback=red!3, colframe=red!50!black, fonttitle=\bfseries\small, title=Generation User Message]
\small
\textit{[If RAG context available:]}\textbackslash n
RETRIEVED CONTEXT (all pre-cutoff):\textbackslash n\texttt{\{context\_block\}}\textbackslash n\textbackslash n
QUESTION: \texttt{\{question\}}\textbackslash n\textbackslash n
BACKGROUND: \texttt{\{background\}}\textbackslash n\textbackslash n
\textit{[If stock task:]}\; TICKERS TO RANK: \texttt{\{tickers\}}
\end{tcolorbox}

\paragraph{Task-Specific Output Schemas.}
Appended to the system prompt via \texttt{\{output\_spec\}}. Each schema enforces structured JSON output with an evidence array, reasoning paragraph, and task-specific prediction field.

\begin{tcolorbox}[colback=red!3, colframe=red!50!black, fonttitle=\bfseries\small, title=Output Schemas (abridged)]
\small
\textbf{Legal:}\textbackslash n
\{``evidence'': [\{``id'': 1, ``fact'': ``...'', ``source\_date'': ``YYYY-MM-DD''\}],\textbackslash n
\phantom{x}``reasoning'': ``...'',\; ``probability\_petitioner'': $\langle$float in [0, 1]$\rangle$\}\textbackslash n\textbackslash n
\textbf{Salary:}\textbackslash n
\{...,\; ``predicted\_salary'': $\langle$positive integer; AAV in USD$\rangle$\}\textbackslash n\textbackslash n
\textbf{Stock:}\textbackslash n
\{...,\; ``ranking'': [``TICKER\_BEST'', ..., ``TICKER\_WORST'']\}\textbackslash n\textbackslash n
\textbf{Binary (large-scale ablation):}\textbackslash n
\{...,\; ``probability\_yes'': $\langle$float 0.0--1.0$\rangle$,\; ``prediction'': ``yes/no''\}
\end{tcolorbox}

\paragraph{RAG Query Generation.}
Used by Temporal RAG and TimeSPEC to generate search queries for the retrieval pipeline. The prompt is user-only (no system message).

\begin{tcolorbox}[colback=red!3, colframe=red!50!black, fonttitle=\bfseries\small, title=RAG Query Generation Prompt]
\small
You are generating web search queries to gather relevant evidence for a forecasting question. All evidence must predate the cutoff date.\textbackslash n\textbackslash n
CUTOFF DATE: \texttt{\{cutoff\_date\}}\textbackslash n\textbackslash n
QUESTION: \texttt{\{question\}}\textbackslash n\textbackslash n
BACKGROUND: \texttt{\{background\}}\textbackslash n\textbackslash n
Generate between \texttt{\{min\_queries\}} and \texttt{\{max\_queries\}} search queries.\textbackslash n
- Each query should target a DISTINCT aspect of the information you need.\textbackslash n
- Include the year in each query to anchor temporal relevance.\textbackslash n
- Do not include queries about events after \texttt{\{cutoff\_date\}}.\textbackslash n\textbackslash n
Output as JSON array: [\{``query'': ``...''\}]
\end{tcolorbox}

\subsection{Supervision and Regeneration Prompts}
\label{app:prompt_sup}

Self-Correction and TimeSPEC share the supervision--regeneration--aggregation loop. The taxonomy verification prompt is also shared with the evaluation pipeline (Section~\ref{app:prompt_eval}).

\paragraph{Taxonomy Verification.}
Used by the generation-time supervisor (Self-Correction, TimeSPEC) and the post-hoc evaluation pipeline. The system prompt defines the auditor role; the user prompt supplies claims and the C0--C8 decision tree.

\begin{tcolorbox}[colback=green!5, colframe=green!50!black, fonttitle=\bfseries\small, title=Taxonomy Verification System Prompt]
\small
You are a temporal provenance auditor.\textbackslash n\textbackslash n
Your task is public-availability date verification, not factual truth checking and not outcome prediction. For each claim, decide the earliest reliable date on which the claim could be publicly verified. Use this date to determine leakage relative to the cutoff. Return only valid JSON matching the requested schema.
\end{tcolorbox}

\begin{tcolorbox}[colback=green!5, colframe=green!50!black, fonttitle=\bfseries\small, title=Taxonomy Verification User Prompt]
\small
Cutoff date: \texttt{\{cutoff\_date\}}\textbackslash n
Task: \texttt{\{task\}}\textbackslash n
Target outcome description: \texttt{\{target\_description\}}\textbackslash n\textbackslash n
Apply the C0-C8 decision tree independently to each item. Stop at the first matching rule.\textbackslash n\textbackslash n
For each item:\textbackslash n
1. Treat declared\_source\_date as a candidate, not ground truth.\textbackslash n
2. Decide whether declared\_source\_date is plausible for when this claim could first be publicly verified.\textbackslash n
3. If implausible, provide the earliest reliable corrected date.\textbackslash n
4. If uncertain, use the conservative latest plausible date and lower confidence.\textbackslash n
5. Do not label a stable/background fact leaked merely because it appears in later litigation.\textbackslash n
6. For claims about a specific case order, holding, filing, grant, denial, oral argument, or opinion, use the public release date.\textbackslash n
7. Leakage: effective\_source\_date $>$ cutoff\_date, except C7/C8 which are leaked by definition.\textbackslash n\textbackslash n
Items: \texttt{\{items\_json\}}\textbackslash n\textbackslash n
Output JSON: \{``results'': [\{``id'': 1, ``category'': ``C0'', ``declared\_date\_plausible'': true, ``effective\_source\_date'': ``YYYY-MM-DD'', ``is\_leaked'': false, ``confidence'': ``high'', ``reason'': ``...''\}]\}
\end{tcolorbox}

\paragraph{Search Re-Verification (TimeSPEC only).}
After the initial taxonomy pass, claims with uncertain or borderline dates are escalated to Perplexity search. The search snippets and original claim are then passed to this prompt for re-verification.

\begin{tcolorbox}[colback=green!5, colframe=green!50!black, fonttitle=\bfseries\small, title=Search Re-Verification Prompt]
\small
\textcolor{green!50!black}{System}: You are a temporal provenance auditor. You have been given a claim and search results that were retrieved to verify the public availability date of that claim.\textbackslash n\textbackslash n
Based on the search results, determine the earliest reliable date on which the information in the claim would first have been publicly knowable.\textbackslash n\textbackslash n
Return only valid JSON: \{``effective\_source\_date'': ``YYYY-MM-DD or null'', ``is\_leaked'': true/false, ``confidence'': ``high$|$medium$|$low'', ``reason'': ``brief explanation''\}

\vspace{0.3em}
\textcolor{green!50!black}{User}: Claim: \texttt{\{claim\_text\}}\textbackslash n
Declared source date: \texttt{\{declared\_date\}}\textbackslash n
Cutoff date: \texttt{\{cutoff\_date\}}\textbackslash n\textbackslash n
Search results:\textbackslash n\texttt{\{search\_snippets\}}\textbackslash n\textbackslash n
Based on these search results, what is the earliest reliable public availability date for this claim? Return JSON only.
\end{tcolorbox}

\paragraph{Regeneration.}
When the supervisor detects leaked claims, the regeneration prompt prepends a leakage warning to the shared generation system prompt. The user message includes the question, optional RAG context, and the previous evidence list for reference.

\begin{tcolorbox}[colback=red!3, colframe=red!50!black, fonttitle=\bfseries\small, title=Regeneration System Prefix (prepended to Generation System Prompt)]
\small
The following evidence items from your previous response were flagged as potentially containing information that was not publicly available before the cutoff date: \texttt{\{leaked\_ids\}}\textbackslash n\textbackslash n
You must generate a new response that does NOT rely on these items. Build a fresh evidence list and reasoning that avoids the leaked information. Your evidence list may include other facts, but not facts that depend on or refer to the flagged information.
\end{tcolorbox}

\paragraph{Aggregation.}
When any leakage was detected across generation cycles, the aggregator synthesizes a final prediction from only the deduplicated validated (non-leaked) evidence. This enforces the closed-world constraint.

\begin{tcolorbox}[colback=red!3, colframe=red!50!black, fonttitle=\bfseries\small, title=Aggregator Prompt]
\small
\textcolor{red}{System}: You are an expert forecaster. You have been given a list of verified evidence items that have been confirmed to predate the knowledge cutoff. Using ONLY this evidence, produce a final structured prediction. Do not introduce any new facts not in the provided evidence list.

\vspace{0.3em}
\textcolor{red}{User}: Evidence items (all verified as pre-cutoff):\textbackslash n\texttt{\{evidence\_json\}}\textbackslash n\textbackslash n
QUESTION: \texttt{\{question\}}\textbackslash n\textbackslash n
BACKGROUND: \texttt{\{background\}}\textbackslash n\textbackslash n
\texttt{\{output\_spec\}}
\end{tcolorbox}

\subsection{Evaluation Pipeline Prompts}
\label{app:prompt_eval}

\paragraph{Claim Extraction (Phase~1).}
Claims are extracted as structured JSON during generation: each method produces an \texttt{evidence} array of \texttt{\{id, fact, source\_date\}} objects as part of its output schema. The evaluation pipeline loads these directly without a separate LLM extraction call. Taxonomy classification uses the verification prompt above (Section~\ref{app:prompt_sup}).

\paragraph{Shapley Coalition Prediction (Phase~2).}
For each coalition subset, the model predicts from only the included claims. The system prompt is stable per instance (cacheable); the user prompt varies per coalition. Task-specific prompts are shown below.

\begin{tcolorbox}[colback=green!5, colframe=green!50!black, fonttitle=\bfseries\small, title=Shapley Coalition --- Legal]
\small
\textcolor{green!50!black}{System}: You are a probability estimator. Output ONLY valid JSON (no markdown, no prose). The JSON object must contain only the key probability\_petitioner --- nothing else.

\vspace{0.3em}
\textcolor{green!50!black}{User}: Predict P(petitioner wins).\textbackslash n
Case: \texttt{\{case\_name\}} | Petitioner: \texttt{\{petitioner\}} | Respondent: \texttt{\{respondent\}}\textbackslash n
Evidence: \texttt{\{evidence\_json\}}\textbackslash n
Reply with exactly: \{``probability\_petitioner'': $\langle$float 0.0-1.0$\rangle$\}
\end{tcolorbox}

\begin{tcolorbox}[colback=green!5, colframe=green!50!black, fonttitle=\bfseries\small, title=Shapley Coalition --- Salary]
\small
\textcolor{green!50!black}{System}: You are an estimator. Output ONLY valid JSON, no other text.

\vspace{0.3em}
\textcolor{green!50!black}{User}: Q: \texttt{\{question\}}\textbackslash n
Facts:\textbackslash n\texttt{\{numbered\_facts\}}\textbackslash n
Output schema: \{``salary'': positive\_number\}
\end{tcolorbox}

\begin{tcolorbox}[colback=green!5, colframe=green!50!black, fonttitle=\bfseries\small, title=Shapley Coalition --- Stock]
\small
\textcolor{green!50!black}{System}: You are a stock analyst. Output ONLY valid JSON (no markdown, no prose). The JSON object must contain only the key ranking --- nothing else.

\vspace{0.3em}
\textcolor{green!50!black}{User}: Rank tickers by expected return.\textbackslash n
Allowed: \texttt{\{ticker\_universe\_json\}}\textbackslash n
Evidence: \texttt{\{evidence\_json\}}\textbackslash n
Reply with exactly: \{``ranking'': [``TICKER1'', ``TICKER2'', ...]\}
\end{tcolorbox}

\begin{tcolorbox}[colback=green!5, colframe=green!50!black, fonttitle=\bfseries\small, title=Shapley Coalition --- Binary]
\small
\textcolor{green!50!black}{System}: You are a probability estimator. Output ONLY valid JSON (no markdown, no prose). The JSON object must contain the single key probability\_yes --- nothing else.

\vspace{0.3em}
\textcolor{green!50!black}{User}: Estimate P(YES) for: \texttt{\{question\}}\textbackslash n
Evidence: \texttt{\{evidence\_json\}}\textbackslash n
Reply with exactly: \{``probability\_yes'': $\langle$float 0.0-1.0$\rangle$\}
\end{tcolorbox}

\paragraph{Leakage Verification (Phase~3).}
For claims in categories C2--C6 that require date verification, the taxonomy verification prompt (Section~\ref{app:prompt_sup}) serves as the primary verification mechanism. Claims with implausible or uncertain dates are escalated to search-based verification via Perplexity API, followed by the search re-verification prompt.